\documentclass{cta-author}

\usepackage{graphicx}
\usepackage{subfig}
\usepackage{url}
\usepackage{booktabs}
\usepackage{multirow}
\usepackage{anyfontsize}
\usepackage[super]{nth}
\usepackage{tikz}
\usetikzlibrary{positioning}

\usepackage{xspace}
\makeatletter
\DeclareRobustCommand\onedot{\futurelet\@let@token\@onedot}
\def\@onedot{\ifx\@let@token.\else.\null\fi\xspace}
\def\etal{\emph{et al}\onedot}

{}
{}
{}

\begin{document}


\title{On the Generalisation Capabilities of Fingerprint Presentation Attack Detection Methods in the Short Wave Infrared Domain}

\author{\au{Jascha Kolberg$^{1\corr}$}, \au{Marta Gomez-Barrero$^{2}$}, \au{Christoph Busch$^{1}$}}

\address{\add{1}{da/sec - Biometrics and Internet Security Research Group, Hochschule Darmstadt, Germany}
\add{2}{Hochschule Ansbach, Germany}
\email{jascha.kolberg@h-da.de}}

\begin{abstract}
Nowadays, fingerprint-based biometric recognition systems are becoming increasingly popular. However, in spite of their numerous advantages, biometric capture devices are usually exposed to the public and thus vulnerable to presentation attacks (PAs). Therefore, presentation attack detection (PAD) methods are of utmost importance in order to distinguish between bona fide and attack presentations. Due to the nearly unlimited possibilities to create new presentation attack instruments (PAIs), unknown attacks are a threat to existing PAD algorithms. This fact motivates research on generalisation capabilities in order to find PAD methods that are resilient to new attacks. In this context, we evaluate the generalisability of multiple PAD algorithms on a dataset of 19,711 bona fide and 4,339 PA samples, including 45 different PAI species. The PAD data is captured in the short wave infrared domain and the results discuss the advantages and drawbacks of this PAD technique regarding unknown attacks.
\end{abstract}

\maketitle
\renewcommand{\thefootnote}{\arabic{footnote}}

\section{Introduction}
\label{sec:intro}
Next to knowledge-based and token-based authentication methods, biometric recognition systems~\cite{Jain-BiometricRecognition-Nature-2007} are established in our daily life. Common examples include high security border control on the one side, and user convenient smartphone unlocking on the other side of the wide range of applications. The unique link between the observed biometric characteristic (i.e., fingerprint) and the data subject's identity are a main advantage of biometric authentication, since passwords and tokens could be shared among individuals. However, biometric systems are usually exposed to the public and thus vulnerable to presentation attacks (PAs). In this case, no bona fide biometric characteristic but an artificial presentation attack instrument (PAI) is presented to the capture device~\cite{ISO-IEC-30107-1-PAD-Framework-160115}.
The goal of the attacker is either to impersonate another data subject or to conceal the own identity due to black-listing. Since successful attacks~\cite{Biggio-EvaluationOfPAs-IETBiometrics-2012, Husseis-SurveyPA+PAD-ICCST-2019} are known, (unsupervised) biometric recognition systems require an automated presentation attack detection (PAD) module in order to detect attack presentations and only process bona fide presentations~\cite{Marcel-HandbookPAD-ACVPR-2019}. Within the last decade, researchers proposed a lot of different PAD approaches~\cite{Sousedik-PAD-Survey-IET-BMT-2014, Marasco-PAD-SurveyFingerprint-CSUR-2014}, that can be classified into two categories: \textit{i) software-based} methods, where the captured data of commercial sensors is analysed in a deeper way; and \textit{ii) hardware-based} methods, where additional PAD data from the biometric characteristic are acquired by an additional sensor integrated in the capture device and then processed with dedicated software.

Fingerprint PAIs can be created from a wide variety of different materials~\cite{Kanich-FingerprintPAIs-IWBF-2018} and in shapes of full fake fingers, thin overlays, or simple printouts. These distinct combinations result in various PAI species with different properties. A continuous and unsolvable challenge is to collect a \lq perfect\rq~dataset including \emph{all} PAI species. Hence, it is very important to evaluate the PAD performance on unknown attacks in order to see the generalisability.

In this context, we apply a generalisation protocol on ten fingerprint PAD algorithms in order to analyse their vulnerability towards unknown attack groups.  The evaluation is carried out on data captured in the short wave infrared domain with over 24,000 samples, including 45 different PAI species. These PAI species are clustered into similarity groups, which are consecutively left out during training of the algorithms and seen only during the testing. 

The remaining article is structured as follows: Section~\ref{sec:sota} summarises related work on generalisation approaches for fingerprint PAD and Section~\ref{sec:sensor} includes the description of the utilised hardware-based capture device. The PAD methods are introduced in Section~\ref{sec:pad} and Section~\ref{sec:eval} defines the experimental setup and protocol. Finally, Section~\ref{sec:results} discusses the results before Section~\ref{sec:conclusions} concludes our findings.

\section{Related Work}
\label{sec:sota}
\begin{table}[t]
	\centering
	\caption{Fingerprint PAD approaches on unknown attacks.}
	\label{tab:sota}
	\setlength{\tabcolsep}{3pt}
	\begin{tabular}{clc}
		\toprule
		Year & Description  & Ref. \\ 
		\midrule
		2010	& Ridge signal, valley noise, region labelling, LOO & \cite{Tan-EffectUnknownPAIs-FingerPAD-WIFS-2010} \\ 
		2011 & Static + intensity-based features, LOO & \cite{Marasco-RobustnessUnknownPAI-FingerPAD-BIOSIGNALS-2011} \\
		2015 & Binary SVM based on One-class SVM, Open set & \cite{Rattani-GeneralisationFingerprintPAD-TIFS-2015} \\
		\midrule
		\multirow{2}{*}{2016} & One-class SVMs + Score fusion & \cite{Ding-OneClassFingerPAD-WIFS-2016} \\
		& CNNs. Unknown PAs, cross sensor/DB & \cite{Nogueira-CNNFingerprintPAD-TIFS-2016} \\
		\midrule
		\multirow{4}{*}{2019} & One-class GANs & \cite{Engelsma-OneClassFingerPAD-ICB-2019} \\
		& Patch-based LSTM + CNN, LOO & \cite{Mirzaalian-LSCI-FingerPAD-2019} \\
		& Fingerprint Spoof Buster, LOO + best subset & \cite{Chugh-PADGeneralisation-ICB-2019, Chugh-FingerprintMaterialGenerator-TIFS-2020} \\
		& Feature encoding, Unknown PAs, cross sensor/DB & \cite{GonzalezSoler-FingerPADFeatEncoding-TBIOM-2019} \\
		\midrule
		\multirow{2}{*}{2020} & CNN, ARL, unknown PAs, cross sensor/DB & \cite{Grosz-CrossSensorFingerprintPAD-Arxiv-2020} \\
		 & One-class convolutional autoencoder & \cite{Kolberg-Autoencoder-FingerPAD-ARXIV-2020}\\
		\bottomrule
	\end{tabular} 
\end{table}

A detailed overview of hardware-based fingerprint PAD is presented in \cite{Kolberg-Autoencoder-FingerPAD-ARXIV-2020}, hence we focus on unknown attacks~\cite{Singh-SurveyUnknownFingerprintPAD-Arxiv-2020} and generalisation approaches in this work. A summary of these works is given in Table~\ref{tab:sota}. It should be noted that, most of the reviewed approaches use different datasets and present multiple experiments including leave-one-out (LOO) or cross database/sensor protocols. For LOO experiments, usually one PAI species is left out from training and only used for testing to evaluate the classification of unknown attacks. Hence, a comparison of observed performance measures is not included due to its lack of fairness.

Starting in 2010, Tan \etal~\cite{Tan-EffectUnknownPAIs-FingerPAD-WIFS-2010} evaluated different environmental conditions and unseen PAIs during testing. In particular, they used the ridge signal, valley noise, and region labelling, and reported much higher error rates for unknown scenarios than for cases where PAIs are available during training. For their experiments, the authors collected a database using three fingerprint capture devices: CrossMatch, Digital Persona, and Identix.
The same devices were also used to collect the LivDet 2009 dataset~\cite{livdet2009}, which was then utilised by Marasco and Sansone~\cite{Marasco-RobustnessUnknownPAI-FingerPAD-BIOSIGNALS-2011} to test further PAD methods. Static and intensity-based features were extracted from the fingerprint samples and classified in a LOO analysis. Through combining multiple PAD methods, the authors reduced the impact of unknown PAIs during training.
Based on the LivDet 2011 dataset~\cite{livdet2011}, Rattani \etal~\cite{Rattani-GeneralisationFingerprintPAD-TIFS-2015} evaluated an open-set scenario on unknown attacks. They create a one-class SVM and fine-tuned it on selected PA samples to produce a binary classifier. This SVM is then used for PAD on partly unknown attacks in the test partition, and is subsequently recalibrated on those new materials.

However, the original LivDet evaluation protocol did not consider unknown attacks at the test stage. LivDet 2015~\cite{livdet2015} was the first dataset including three additional unknown attack instruments in the test set. The subsequent competitions, LivDet 2017~\cite{livdet2017} and LivDet 2019~\cite{livdet2019}, have only unknown attacks in the test set and are completely trained on different PAI species.

Recently, the generalisation capabilities of PAD algorithms attracted more attention. In this context, Chugh and Jain~\cite{Chugh-PADGeneralisation-ICB-2019}, performed a LOO analysis of their Fingerprint Spoof Buster~\cite{Chugh-FingerprintSpoofBuster-TIFS-2018} on a combined dataset from MSU-FPAD and PBSKD. Furthermore, they defined a generalisation subset, which includes only six out of twelve PAI species, but enables the detection of the unknown materials due to their similarity. The authors extended this work in \cite{Chugh-FingerprintMaterialGenerator-TIFS-2020} and added the LivDet 2017 dataset as well as a proposed universal material generator to create additional synthetic samples for training.

Furthermore, Nogueira \etal~\cite{Nogueira-CNNFingerprintPAD-TIFS-2016} report increasing error rates for their CNNs on the LivDet 2011~\cite{livdet2011}, LivDet 2013~\cite{livdet2013}, and LivDet 2015~\cite{livdet2015} datasets when introducing unknown scenarios. Their experiments include unknown attacks, cross database, and cross sensor protocols.
Another generalisation approach was presented by Gonzalez-Soler \etal~\cite{GonzalezSoler-FingerPADFeatEncoding-TBIOM-2019} based on the same three LivDet datasets. The authors propose a combination of local features (scale-invariant feature transform) and three feature encodings: bag of words, Fisher vector, and vector locally aggregated descriptors. Those methods are evaluated on unknown attacks as well as cross-database and cross-sensor scenarios. The results show that the Fisher vector encoding performs best on the different settings. 
More recently, Grosz \etal~\cite{Grosz-CrossSensorFingerprintPAD-Arxiv-2020} applied adversarial representation learning (ARL) for fingerprint PAD on the LivDet 2015~\cite{livdet2015}, LivDet 2017~\cite{livdet2017}, and the MSU-FPAD~\cite{Chugh-FingerprintSpoofBuster-TIFS-2018} datasets. They utilise the universal material generator from \cite{Chugh-FingerprintMaterialGenerator-TIFS-2020} in combination with ARL to train a base CNN and achieve domain adaptation and domain generalisation. ARL is especially suited for cross sensor experiments and unknown PAs since it requires no information about the unseen target domain. Their evaluation includes cross sensor, cross DB, and LOO analyses, which results in an improved PAD performance compared to previous approaches.

Following the idea of LOO analysis, Mirzaalian \etal~\cite{Mirzaalian-LSCI-FingerPAD-2019} worked on temporal sequences of laser illuminated images from a private dataset. As classifiers they utilise convolutional neural networks (CNNs) and a long short-term memory (LSTM) network. The latter one can directly process temporal information within sequences while classical CNNs are applied on static images. Their results show a slightly better performance for the LSTM.
Further PAD approaches consider all PAI species as unknown attacks and train one-class classifiers only on bona fide samples. The test sample is then processed in the same way and the classifier validates whether the current sample is similar enough to the ones seen during training. The idea is that PA samples differ from bona fide ones and thus can be detected.
Using one-class support vector machines (SVMs), Ding and Ross~\cite{Ding-OneClassFingerPAD-WIFS-2016} trained on twelve different feature sets of the LivDet 2011 database~\cite{livdet2011}. A subsequent score fusion counters the weaknesses of single SVMs and provides more generalisability.
Engelsma and Jain~\cite{Engelsma-OneClassFingerPAD-ICB-2019} used three different generative adversarial networks (GANs) on an own collected dataset. Their one-class approach is based on the DCGAN architecture proposed by Radford \etal~\cite{DCGANarchitecture}. 
Finally, Kolberg \etal~\cite{Kolberg-Autoencoder-FingerPAD-ARXIV-2020} built a convolutional autoencoder which was trained on bona fide samples captured in the short wave infrared (SWIR) domain between 1200 nm and 1550 nm. 
Their approach achieved superior detection performance compared to other one-class classifiers such as SVMs or Gaussian mixture models.

\section{Capture Device and PAD data}
\label{sec:sensor}
\begin{figure}[t]
	\centering
	\includegraphics[width=\linewidth]{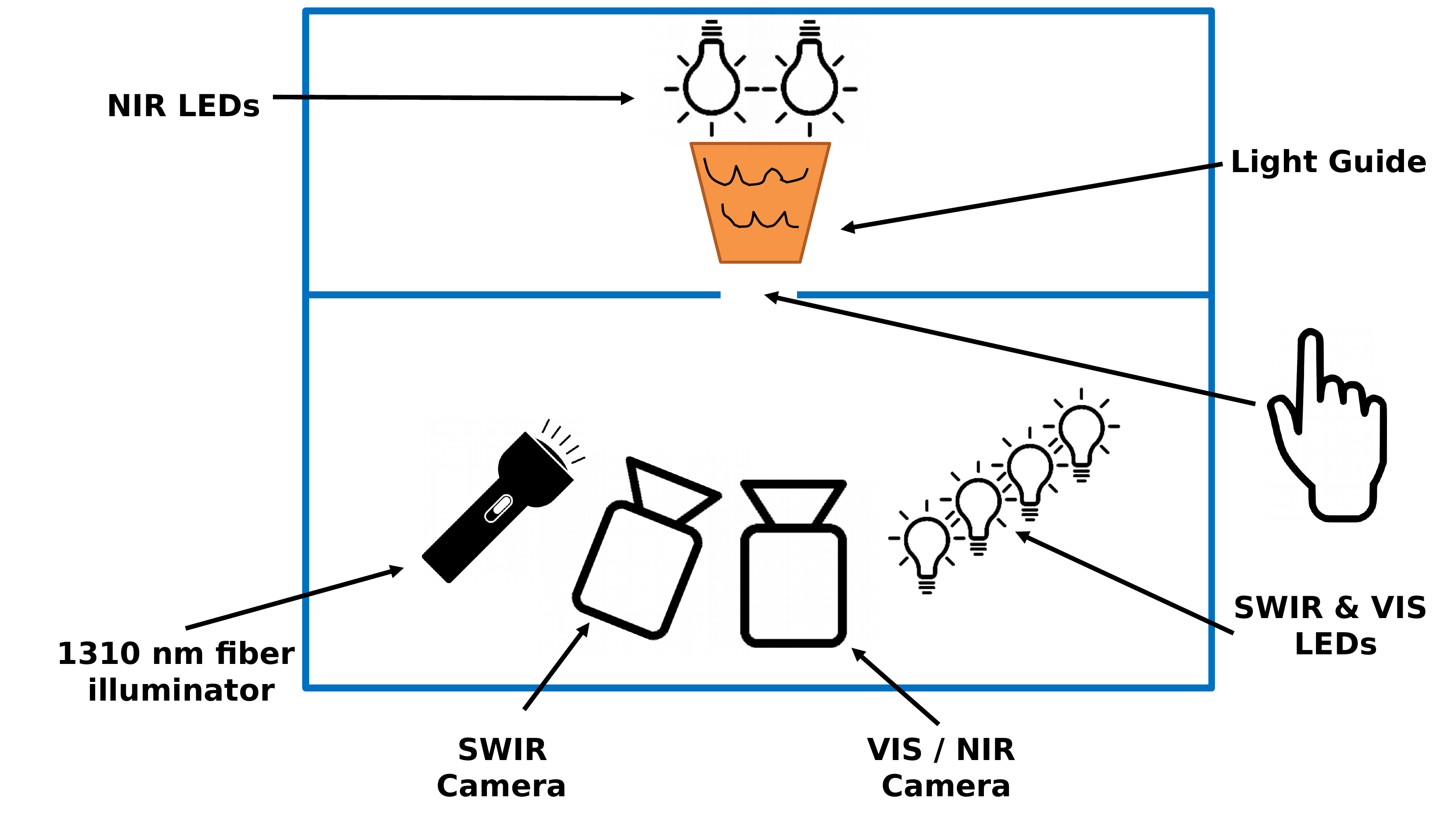}
	\caption{The capture device is a closed box with only one free slot for the finger. Two cameras in combination with multiple illuminations are able to capture the fingerprint and additional PAD data.}
	\label{fig:sensor}
\end{figure}
\begin{figure}[t]
	\centering
	\subfloat[][]{\includegraphics[width=0.3\linewidth]{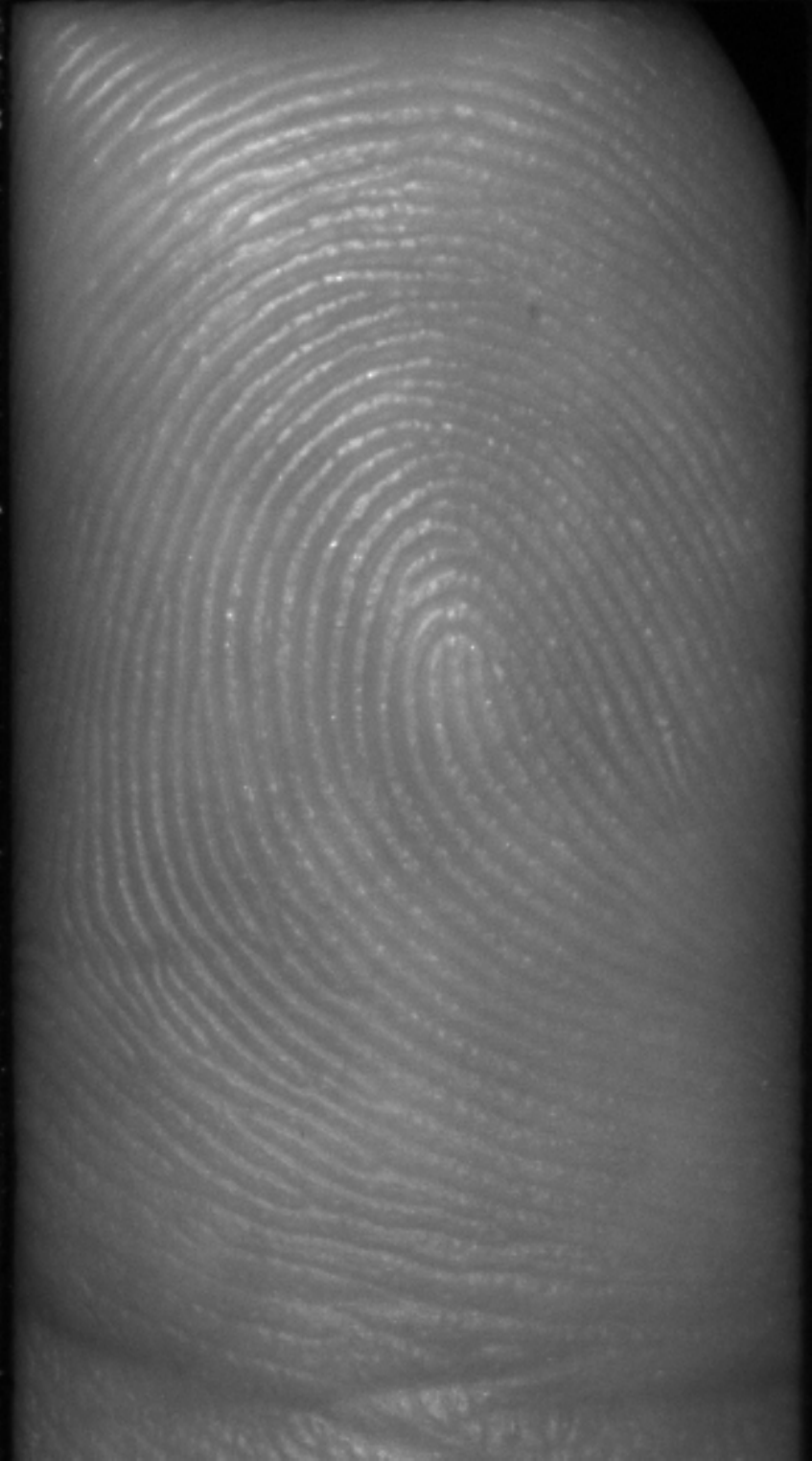}}
	\subfloat[][]{\includegraphics[width=0.3\linewidth]{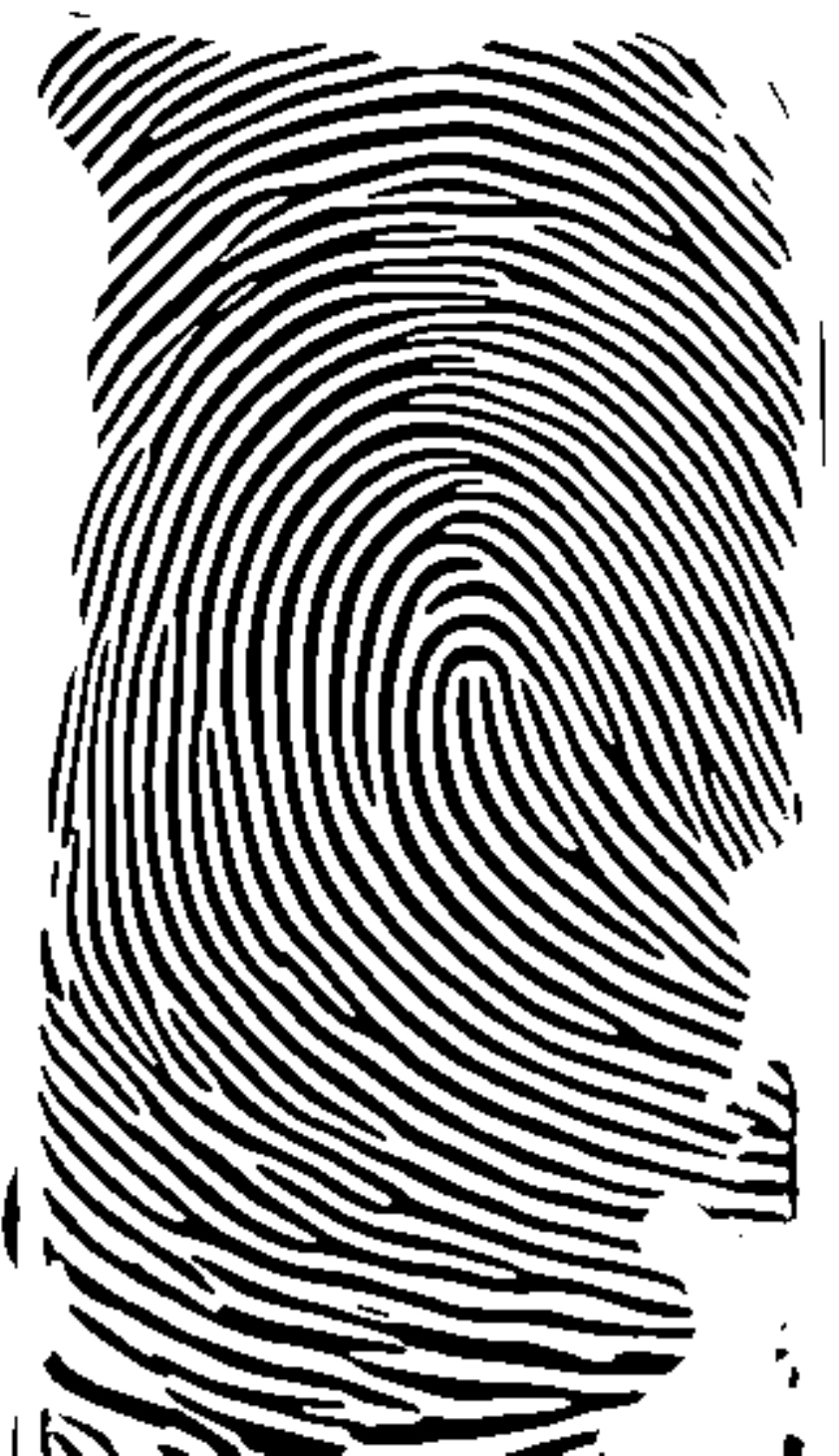}}
	\subfloat[][]{\includegraphics[width=0.3\linewidth]{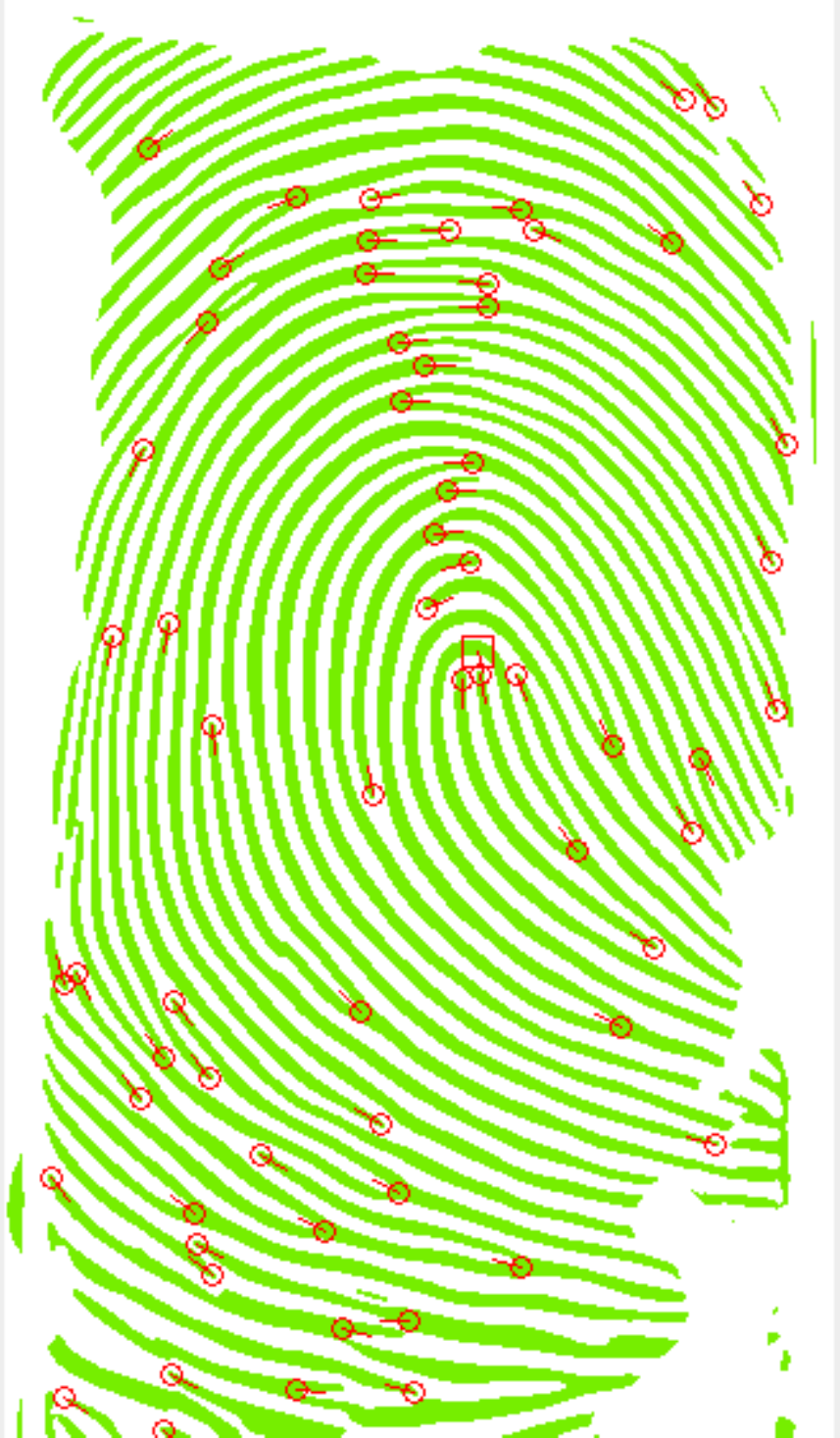}}
	\caption{Fingerprint recognition pipeline for legacy compatibility: (a) original finger photo, (b) extracted fingerprint, and (c) detected minutiae points.}
	\label{fig:fp_recognition}
\end{figure}
For the data collection, a camera-based fingerprint capture device\footnote{An in-depth description of the capture device is given in \cite{Spinoulas-BATLDatasetFingerPAD-Arxiv-2020, Spinoulas-BATLDesignCaptureDevices-Arxiv-2020}.} was used as depicted in Fig.~\ref{fig:sensor}. One camera (Basler acA1300-60gm) takes finger photos in the visible spectrum to extract the fingerprint for legacy compatibility. This process is illustrated in Fig.~\ref{fig:fp_recognition}, where Neurotechnology Verifinger SDK\footnote{\url{https://www.neurotechnology.com/verifinger.html}} was used for steps (b) and (c). Additional finger vein images can be captured when activating the near-infrared (NIR) LEDs above the finger. A second camera (100 fps Xenics Bobcat 320) captures PAD data in wavelengths between 900 nm and 1700 nm. Both cameras are placed in a closed box next to multiple illumination sources with only one finger slot at the top. Once a finger is placed on this slot, all ambient light is blocked and only the desired wavelengths illuminate the finger. The invisible SWIR wavelengths of 1200 nm, 1300 nm, 1450 nm, and 1550 nm are especially suited for PAD because all skin types in the Fritzpatrick scale~\cite{Fritzpatrick-SkinTypes-Dermatology-1988} reflect in the same way as shown by Steiner \etal~\cite{Steiner-facePADswir-ICB-2016} for face PAD, but on the contrary PAI species reflect quite different from skin. Hence, SWIR images are captured in each of these wavelengths.
Additionally, a 1310 nm laser diode illuminates the finger area and a sequence of 100 frames is collected within one second. Stemming from biomedical applications, this laser sequence is used to image and monitor microvascular blood flow~\cite{Senarathna-LSCI-IEEERevBiomedEng-2013}. Since the laser scatters differently when penetrating human skin in contrast to artificial PAIs, this technique qualifies for PAD as well as shown in \cite{Keilbach-Fingerprint-LSCI-PAD-BIOSIG-2018, Kolberg-LSCIBenchmarkFingerPAD-BIOSIG-2019}.

\begin{figure}[t]
	\centering
	\input{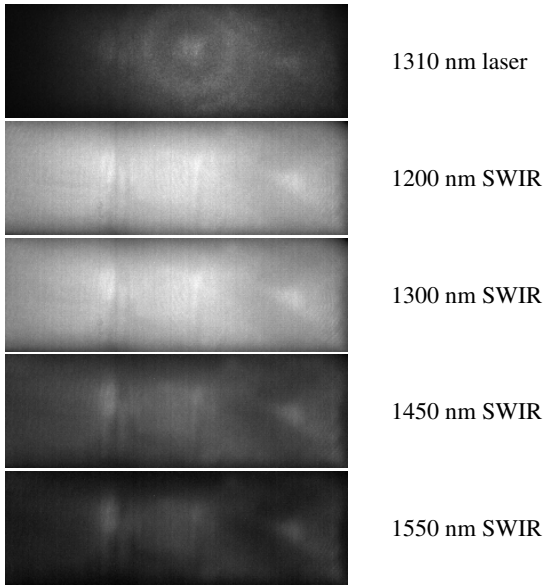}
	\caption{Bona fide samples acquired at five different wavelengths.}
	\label{fig:bfsamples}
\end{figure}
\begin{figure}[t]
	\centering
	\input{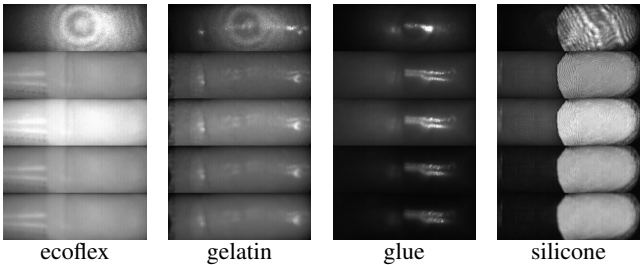}
	\caption{Samples of four different PAs accross all five wavelengths.}
	\label{fig:pasamples}
	\vspace*{-2.em}
\end{figure}
The total capture time for all these data takes two to four seconds with this prototype.
Example frames of a bona fide presentation acquired at the aforementioned wavelengths are shown in Fig.~\ref{fig:bfsamples}. For the laser sequence data, only one frame is depicted since the subtle temporal changes are not visible in steady pictures. Nevertheless, we can recognise a circle where the laser focuses the finger. On the other hand, the LEDs achieve a much more consistent illumination for the SWIR images, where the skin reflections get darker for increasing wavelengths. For comparison, Fig.~\ref{fig:pasamples} shows PA samples made from ecoflex, gelatin, glue, and silicone. Depending on the PAI material and type, the images differ more or less from the bona fide images. Especially, the silicone sample responds differently on the overlay part in contrast to the skin part of the same presentation.
The region of interest for all samples comprises $100\times 300$ pixels due to the fixed size of the finger slot.

\section{Presentation Attack Detection}
\label{sec:pad}
\begin{figure*}[t]
	\centering
	\input{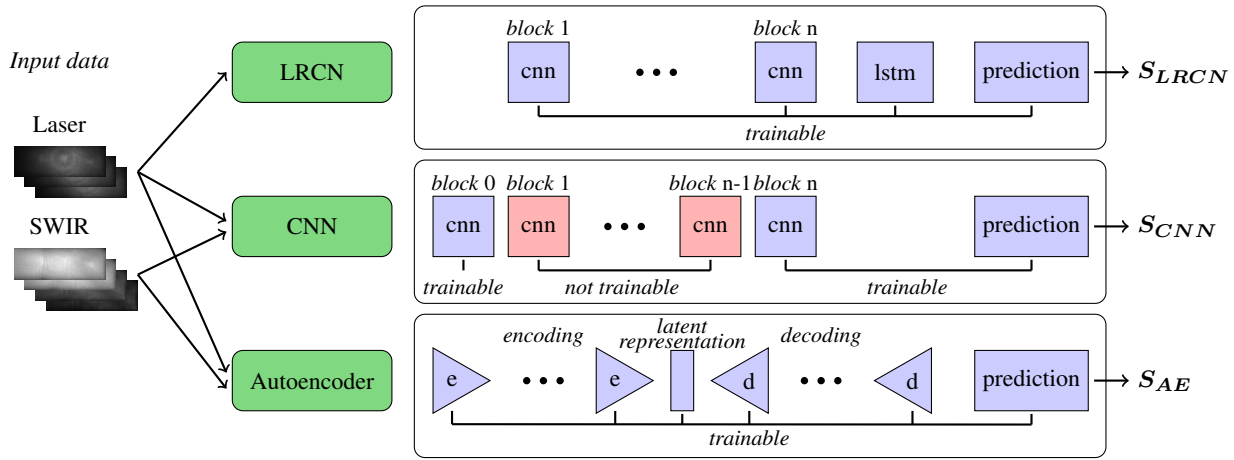}
	\caption{Scheme of the PAD methods analysed. The LRCN receives a Laser sequence of 100 frames. The remaining CNNs and Autoencoder receive a 3-dimensional input from selected Laser frames or a 4-dimensional SWIR input. On the right we can observe more details about the networks' architecture: Each CNN is split into $n$ blocks. In addition, block 0 applies only for SWIR input data in order to automatically reduce the 4-dimensional input to a 3-dimensional one as proposed in \cite{GomezBarrero-MSSWIRCNN-CRC-2020}. Blue colour indicates trainable blocks, while blocks that are not trainable are shown in red. Each algorithm produces its own detection score in the end.}
	\label{fig:scheme}
\end{figure*}

Based on previously published benchmarks~\cite{Kolberg-Autoencoder-FingerPAD-ARXIV-2020, Kolberg-LSTM-FingerPAD-IJCB-2020, GomezBarrero-MSSWIRCNN-CRC-2020}, we select the best-performing algorithms in order to evaluate their generalisation capabilities towards unknown attacks. For this purpose, three different classes of PAD algorithms are utilised, which are summarised in Fig.~\ref{fig:scheme} and introduced in the following subsections. Each algorithm produces its own detection score, thereby allowing a subsequent score-fusion.

\subsection{Selected Laser algorithms}
Given the laser sequence data, which comprises 100 frames captured at 1310 nm, \cite{Kolberg-LSTM-FingerPAD-IJCB-2020} evaluated PAD algorithms based on temporal and/or spatial features. The LSTM~\cite{lstm} is designed to learn long-term temporal dependencies by maintaining a state of the previously seen data. However, it is not designed to process multi-dimensional input and hence is placed on top of a CNN. The CNN structure serves as a feature extraction in order to reduce the dimension of the input data for the LSTM. In case of utilising a pre-trained CNN base, the PAD performance was not sufficient. However, the long-term recurrent convolutional network (LRCN)~\cite{lrcn} approach achieved the best results. This recurrent sequence model is directly connected to a visual CNN and jointly trained from scratch to learn temporal dynamics and convolutional perceptual representations together (Fig.~\ref{fig:scheme}, top). Apart from the LRCN, which is based on VGG16~\cite{vgg}, three additional CNN methods achieved good results. These CNNs receive a 3-dimensional input created from three specific laser frames since literally no spacial changes are visible within the one second sequence. Thus, using all 100 frames in the CNN would more likely result in over-fitting instead of enhancing the detection accuracy. From the pre-trained VGG16~\cite{vgg} and the VGGFace~\cite{vggface} networks only the last block as well as the prediction layer are re-trained on the PAD data (Fig.~\ref{fig:scheme}, mid). While VGG16 was trained on ImageNet~\cite{imagenet}, VGGFace has seen much more skin during training as the labeled faces in the wild database~\cite{lfw} was used. Furthermore, the small 5-layer residual network (ResNet) proposed in \cite{GomezBarrero-PAD-SWIR-LSCI-ICB-2019} is trained from scratch on the given dataset. Its residual connection consists of reinjecting previous representations into its downstream flow by adding prior tensors to a later one. This structure significantly decreases training time~\cite{He-ResidualDeepLearning-CVPR-2016, Szegedy-InceptionV4-AI-2017}, while preventing information loss during the processing.

\subsection{Selected SWIR algorithms}
Since classical CNNs require a 3-dimensional (RGB) image as input and we capture four different SWIR wavelengths, our previous work~\cite{Tolosana-SWIR-PAD-CNNs-TIFS-2020} proposed a manual dimension reduction given a fixed formula. Hence, all four wavelengths were used to obtain a RGB image. However, this approach is based on a much smaller dataset captured with a different camera (fingerphoto of $18\times 58$ pixels). 
After the hardware change, the performance of that method was clearly outperformed by \cite{GomezBarrero-MSSWIRCNN-CRC-2020}, which defined a multi-spectral pre-processing block that is added in front of the desired CNN. This block 0 (Fig.~\ref{fig:scheme}, mid) receives a 4-dimensional image combined from the four SWIR wavelengths and outputs a 3-dimensional image. It is trained together with the last block and the prediction layer in order to automatically find the best-suited transformation. The resulting 3-dimensional output is then fed to the traditional CNNs. From these SWIR PAD algorithms, the three VGG-based networks (VGGFace~\cite{vggface}, VGG16 and VGG19~\cite{vgg}) achieved the best results as well as the MobileNetV2~\cite{mobilenetv2}. The latter one also makes use of residual connections and additional inverted bottlenecks, where previous bottlenecks are residually connected to subsequent ones. Moreover, given the depth of MobileNetV2, only twelve out of 16 blocks are used in order to adjust for the relative small training set in contrast to other deep learning tasks. In general, all four SWIR CNNs apply transfer learning based on a pre-trained network.

\subsection{One-class Autoencoder}
All previously named PAD methods are two-class algorithms. However, especially for unknown attack evaluations it is interesting to additionally include one-class classifiers in the experiments. Hence, a convolutional autoencoder~\cite{Kolberg-Autoencoder-FingerPAD-ARXIV-2020} (Fig.~\ref{fig:scheme}, bottom) is analysed as well. It works on 3-dimensional laser images and 4-dimensional SWIR images likewise. The encoding phase repeatedly reduces the input dimension until it results in a 1-dimensional latent representation of fixed size. Subsequently, the decoding phase reconstructs the original image size. The final prediction is done by computing the difference between the input and output images. Since the autoencoder is only trained on bona fide samples, it is expected that the reconstruction of PA samples significantly differs from its input and thus can be detected. Previous tests~\cite{Kolberg-Autoencoder-FingerPAD-ARXIV-2020} have shown that a fusion of laser and SWIR autoencoders does not improve the PAD accuracy since the SWIR autoencoder on its own performs better. However, we include both algorithms in this new benchmark for completeness.

\subsection{Fusion}
In contrast to the autoencoder results~\cite{Kolberg-Autoencoder-FingerPAD-ARXIV-2020}, other approaches~\cite{Gomez-Barrero-FusionBATL-PAD-UBIO-2018, Hussein-LSCI-SWIR-CNN-FingerPAD-WIFS-2018, GomezBarrero-PAD-SWIR-LSCI-ICB-2019, Spinoulas-BATLDatasetFingerPAD-Arxiv-2020} have shown improvements for a fusion of laser and SWIR algorithms. However, since the main part of this work consists of analysing the generalisation capabilities, we compute a fixed fusion and evaluate its performance next to the single algorithms on unknown attacks. In this regard, we combine one laser algorithm with one SWIR algorithm in order to observe to what extent this fusion generalises better across different unknown attack species.

\section{Experimental Evaluation}
\label{sec:eval}
Given the different PAD methods, this Section provides further details on the experimental setup and the different evaluated protocols. 
\subsection{Experimental Setup}
\begin{table}[t]
	\centering
	\caption{Summary of PAIs in the database with their corresponding group. The \hspace{\textwidth} numnumber of total samples and the number of variations is given. Variations include \hspace{\textwidth} e.g.\ e.g.\ different colours and conductive augmentations.}
	\label{tab:pais}
	\begin{tabular}{llcc}
		\toprule
		\textbf{PAI Group} & \textbf{PAI} & \textbf{\# variations} & \textbf{\# samples} \\ 
		\midrule
		\multirow{8}{*}{Fakefinger} & 3D printed & 2 & 72 \\ 
		& dental material & 1 & 33 \\ 
		& dragon skin & 3 & 477 \\ 
		& ecoflex & 4 & 291 \\ 
		& latex & 2 & 147 \\ 
		& playdoh & 4 & 116 \\ 
		& silly putty & 3 & 55 \\ 
		& wax & 1 & 74 \\ 
		\midrule
		\multirow{8}{*}{Overlay opaque} & bandage plaster & 1 & 14 \\ 
		& dental material & 1 & 51 \\ 
		& dragon skin & 1 & 17 \\ 
		& ecoflex & 2 & 1035 \\ 
		& gelatin & 1 & 194 \\ 
		& printout paper & 1 & 49 \\
		& silicone & 4 & 752 \\ 
		& urethane & 1 & 72 \\ 
		\midrule
		\multirow{7}{*}{Overlay transparent} & dragon skin & 1 & 106 \\ 
		& gelatin & 1 & 107 \\ 
		& glue & 2 & 27 \\ 
		& latex & 1 & 34 \\ 
		& printout foil & 1 & 64 \\
		& silicone & 1 & 157 \\ 
		& wax & 1 & 18 \\ 
		\midrule
		\multirow{4}{*}{Overlay semi} & dragon skin & 1 & 47 \\ 
		& ecoflex & 1 & 24 \\ 
		& glue & 2 & 146 \\ 
		& silicone & 1 & 160 \\ 
		\midrule
		\midrule
		Mat. group i) & silicone & 7 & 1141 \\
		\midrule
		\multirow{2}{*}{Mat. group ii)} & dragon skin & 6 & 647 \\
		& ecoflex & 7 & 1350 \\
		\midrule
		\multirow{5}{*}{Mat. group iii)} & gelatin & 2 & 301 \\
		& glue & 4 & 173 \\
		& latex & 3 & 181 \\
		& printout & 2 & 113 \\
		& wax & 2 & 92 \\
		\midrule
		\multirow{4}{*}{Mat. group iv)} & 3D printed & 2 & 72 \\
		& dental material & 2 & 84 \\
		& playdoh & 4 & 116 \\
		& silly putty & 3 & 55 \\
		\bottomrule
	\end{tabular} 
\end{table}

The data was collected in four acquisition sessions in two distinct locations within a timeframe of nine months. Subjects could participate multiple times and presented six to eight fingers per capture round including thumb, index, middle, and ring fingers.
Fingers were presented as they were, thereby resulting in different levels of moisture, dirt, or ink. 
The combined database contains a total of 24,050 samples comprising 19,711 bona fides and 4,339 PAs stemming from 45 different PAI species. These PAI species were selected by the project sponsor and include full fake fingers and more challenging overlays as summarised in Table~\ref{tab:pais}. The printouts were also worn as overlays and conductive paint was applied to some PAIs.
With 45 PAI species, leaving out a single PAI species or material at a time would result in a lot of experiments with limited statistical meaning. Hence, we are leaving out complete PAI groups for more relevant results. 
In particular, we defined four relevant PAI groups based on their visual properties: \emph{Fakefinger, Overlay opaque, Overlay transparent, Overlay semi}. For non camera-based systems, other groups (e.g.\ describing the moisture level) would be more relevant than the colour or transparency level. 
In addition to those property groups, we created four material groups, where selected materials are completely excluded from the training process. Given the different numbers of samples per material, we grouped similar materials as follow: \emph{group~i)} silicone; \emph{group~ii)} dragon skin and ecoflex; \emph{group~iii)} gelatine, glue, latex, printout, and wax; \emph{group~iv)} 3D printed, dental material, playdoh, and silly putty.

Note that the owner of the data has indicated a commitment to make the complete dataset available in the near future for reproducibility and benchmarking\footnote{\url{https://www.isi.edu/projects/batl/data}}.

\subsection{Experimental Protocol}
In order to evaluate the generalisation capabilities of the selected PAD algorithms, a leave-one-out protocol is adopted. In particular, the protocol leaves a complete PAI group out of training, which is only used during testing. In addition to the four presented groups in Table~\ref{tab:pais} another group comprises all overlay PAIs, thus training only on fake fingers.

\begin{table}[t]
	\centering
	\setlength{\tabcolsep}{7pt}
	\caption{Number of samples within the different partitions.}
	\label{tab:partition}
	\begin{tabular}{lccc}
		\toprule
		& Training & Validation & Test \\ 
		\midrule
		Baseline (BF) & 807 & 542 & 16,381 \\ 
		Baseline (PA) & 807 & 542 & 2990 \\
		\midrule
		LOO (BF) & 9956 & 3069 & 6686 \\
		Fakefinger & 2624 & 450 & 1265 \\
		Overlay & 1027 & 238 & 3074 \\
		Opaque & 1801 & 354 & 2184 \\
		Transparent & 3152 & 674 & 513 \\
		Semi & 3299 & 663 & 377 \\
		Mat. group i) & 2657 &	541 & 1141 \\
		Mat. group ii) & 2023 & 319 & 1997 \\
		Mat. group iii) & 2884 & 595 & 860 \\
		Mat. group iv) & 3364 & 648 & 327 \\
		\bottomrule
	\end{tabular} 
\vspace*{-1em}
\end{table}
As a baseline, we utilise the partitioning from \cite{Kolberg-LSTM-FingerPAD-IJCB-2020}, where each PAI species is present in training, validation, and test sets. Furthermore, some bona fide samples have been removed in order to grant unbiased training and validation sets with an equal number of PA and bona fide samples, while no data subject is seen during training and testing procedures.
For the LOO partitions, one group of PAIs is only present in the test set. The remaining PA samples are randomly assigned to the training (85\%) and validation (15\%) sets. In order to be able to analyse the influence of different PAI groups during the training procedure, the bona fide samples in each set are identical across all LOO partitions; 50\% for training, 15\% for validation, and 35\% for testing. The specific number of samples in each partition are given in Table~\ref{tab:partition}.
The system requirements depend on the utilised network itself and the particular size of training and validation partitions, since these samples have to be loaded into memory. Thus, our laser CNNs need approximately 8 GB of memory, SWIR CNNs 32 GB, and the laser LRCN 150 GB, respectively. As a consequence, training is considered expensive in terms of resources and time needed. However, training is done only once, possibly offline, and the prediction is done in real time based on the samples of one capture process.

The PAD performance is shown in detection error tradeoff (DET) plots and evaluated according to the standard ISO/IEC 30107-3 on biometric presentation attack detection - Part 3: Testing and Reporting~\cite{ISO-IEC-30107-3-PAD-metrics-170227}. 
To that end, two metrics are used:\\
\textbf{Attack Presentation Classification Error Rate (APCER)}: proportion of attack presentations incorrectly classified as bona fide presentations.\\
\textbf{Bona fide Presentation Classification Error Rate (BPCER)}: proportion of bona fide presentations incorrectly classified as attack presentations.

It should be noted that the PAD threshold can be adjusted depending on the use case. In general, a low BPCER represents a very convenient system, while a low APCER is more important for high security applications. Since we analyse the generalisability on LOO partitions, we compare the algorithms at the single operation point APCER$_{0.2}$. This represents the APCER for a fixed BPCER of 0.2\%.

\section{Experimental Results}
\label{sec:results}
The result analysis follows the structure of the experimental protocol. In order to evaluate the different LOO groups, first the results of the baseline partition are viewed. Subsequently, visual as well as material LOO experiments are discussed.
\subsection{Baseline Performance}

\begin{figure}[t]
	\centering
	\includegraphics[width=0.87\linewidth]{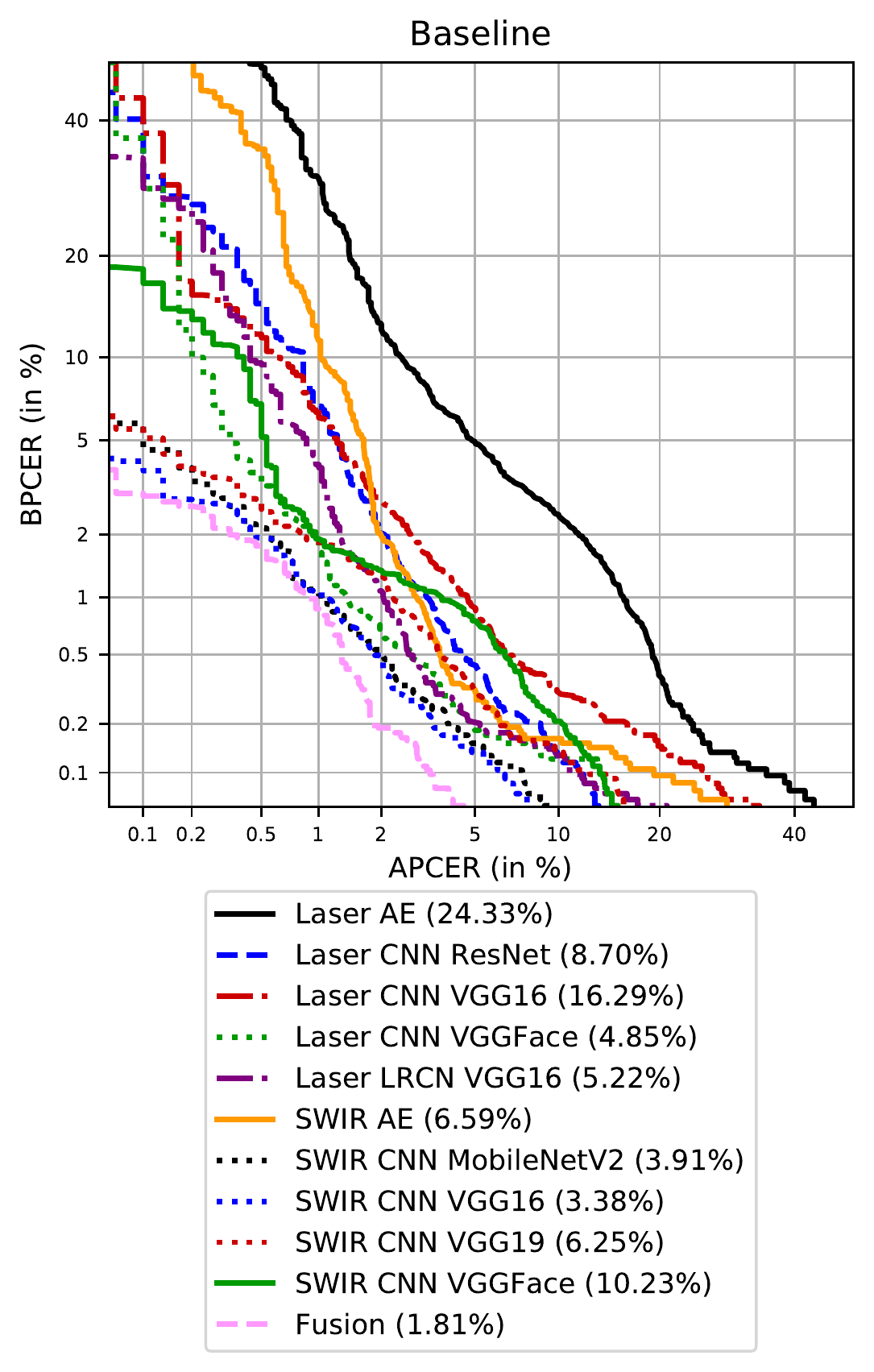}
	\caption{DET curves based on the mixed training partition of \cite{Kolberg-LSTM-FingerPAD-IJCB-2020}. APCER$_{0.2}$ values are given in brackets.}
	\label{fig:det_baseline}
\end{figure}
\begin{figure}[t]
	\centering
	\includegraphics[width=0.97\linewidth]{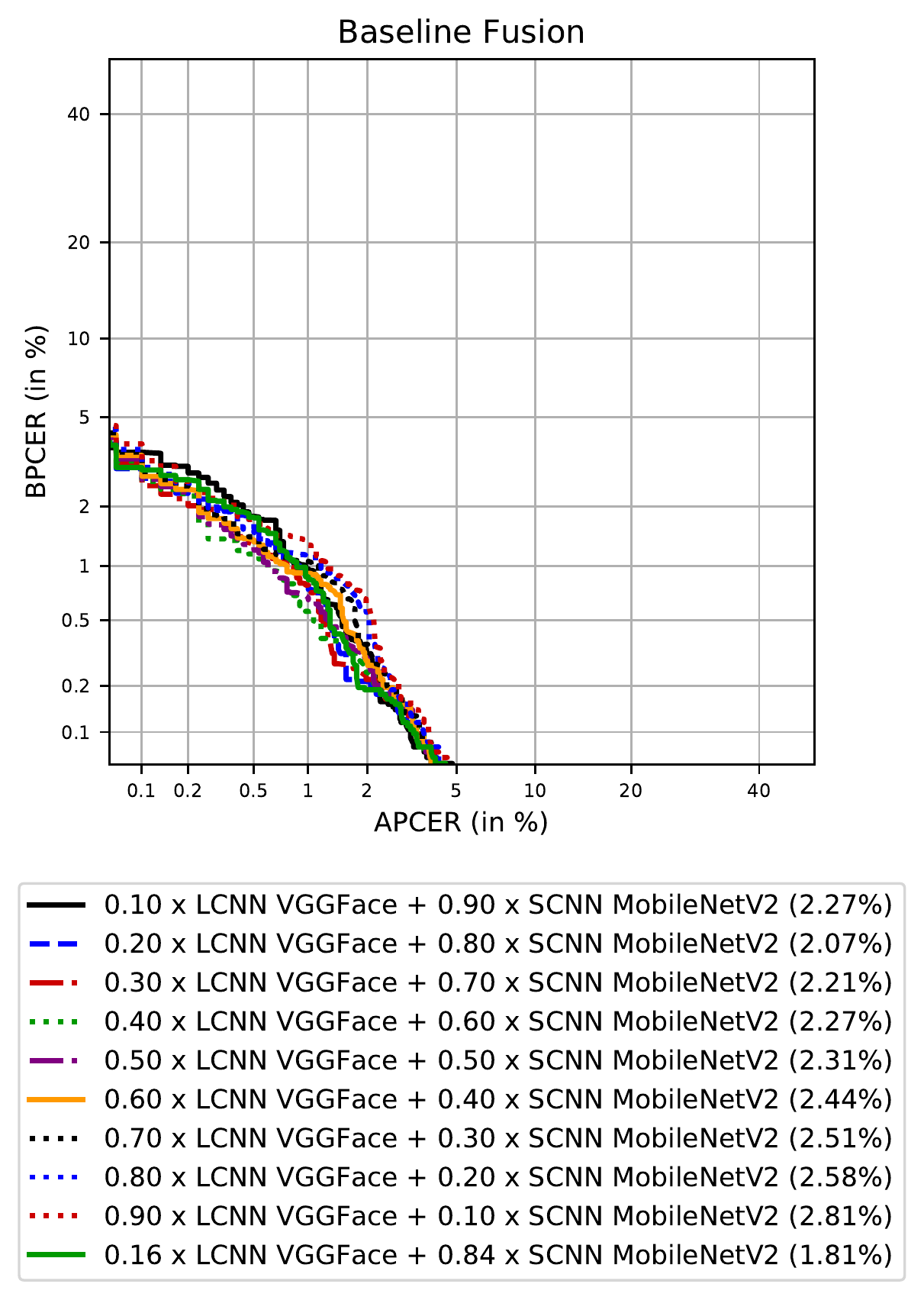}
	\caption{DET curves of different weighted fusions of laser CNN VGGFace and SWIR CNN MobileNetV2. APCER$_{0.2}$ values are given in brackets.}
	\label{fig:det_fusions}
\end{figure}

The first DET plot in Fig.~\ref{fig:det_baseline} shows the algorithms' performance on the baseline partition~\cite{Kolberg-LSTM-FingerPAD-IJCB-2020}. The best results in terms of low APCER$_{0.2}$ values are achieved by the SWIR CNNs VGG16 and MobileNetV2, followed by laser CNN VGGFace and laser LRCN VGG16. Hence, we analysed whether these algorithms complement each other and counted the number of identical 
attack presentation classification errors (APCEs).
Table~\ref{tab:identical_apces} shows how many identical PAI samples are wrongly classified by the algorithms. Since the combination of laser CNN VGGFace and SWIR CNN MobileNetV2 has the least identical APCEs, we tried different weighted fusions of these two algorithms. Fig.~\ref{fig:det_fusions} provides an overview of the influence of different fusion weights. The evaluation showed that performances are quite close and the most generalisable fusion can be achieved by using the following weights of Eq.~\eqref{eq:fusion}:
\begin{equation}
\label{eq:fusion}
0.16 \times S^{Laser}_{CNN-VGGFace} + 0.84 \times S^{SWIR}_{CNN-MobileNetV2}
\end{equation} 
In general, it makes sense to fuse algorithms which do not have much APCEs in common, since such a fusion takes advantage of the strengths of different algorithms.
This combination of laser and SWIR approaches improves the PAD performance to an APCER$_{0.2}$ of 1.81\% on the baseline partition.
Furthermore, the number of APCEs per PAI group for the four best algorithms and the fusion are given in Table~\ref{tab:apces_baseline}. 
Finally, this fusion of the two algorithms and their weights is now fixed for further LOO experiments to observe the generalisability of single algorithms compared to a pre-defined fusion. 

\begin{table}[t]
\centering
\setlength{\tabcolsep}{5pt}
\caption{Number of identical APCEs for different PAD algorithms.}
\label{tab:identical_apces}
\begin{tabular}{lcc}
	\toprule
	& SWIR MobileNetV2 & SWIR VGG16\\
	\midrule
	Laser CNN VGGFace & 23 (0.77\%) & 26 (0.87\%)\\
	Laser LRCN VGG16 & 46 (1.54\%) & 49 (1.64\%)\\
	\bottomrule
\end{tabular}
\end{table}
\begin{table}[t]
\centering
\setlength{\tabcolsep}{3pt}
\caption{Number of APCEs at an APCER$_{0.2}$ on the baseline partition.}
\label{tab:apces_baseline}
\begin{tabular}{lccccc}
\toprule
\multirow{2}{*}{PAI group} & \multicolumn{2}{c}{Laser} & \multicolumn{2}{c}{SWIR} & \multirow{1}{*}{Fusion} \\
 & LRCN & VGGFace & MobileNetV2 & VGG16 & \\
\midrule
Fakefinger & 64 & 57 & 56 & 58 & 26 \\
Opaque & 23 & 23 & 12 & 14 & 2 \\
Transparent & 61 & 46 & 45 & 26 & 26 \\
Semi & 8 & 19 & 4 & 3 & 0 \\
\midrule
total & 156 & 145 & 117 & 101 & 54 \\
\bottomrule
\end{tabular}
\end{table}

\subsection{Visual LOO Groups}
\begin{figure}[t]
	\centering
	\includegraphics[width=0.87\linewidth]{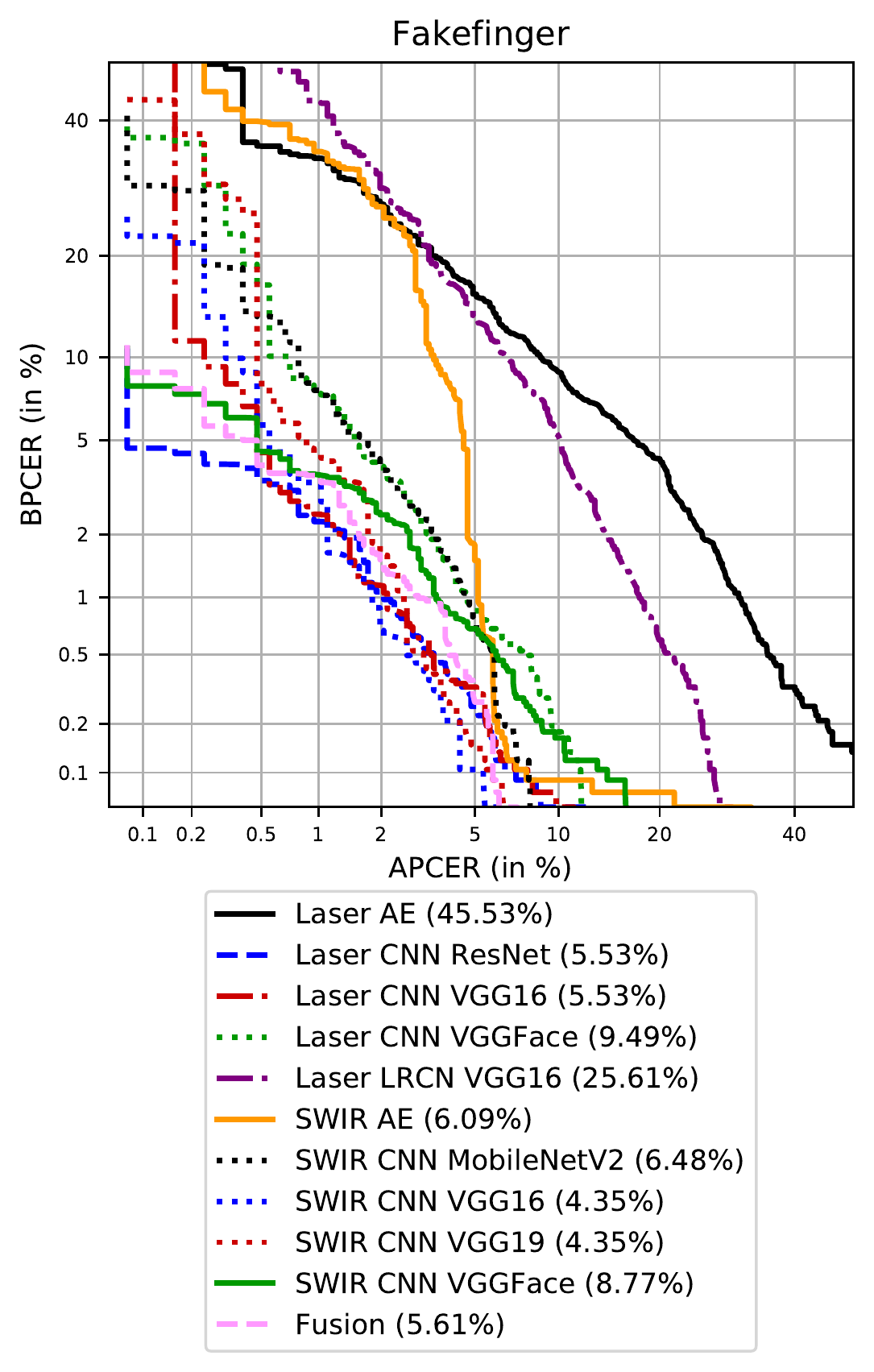}
	\caption{DET curves of the \emph{Fakefinger} group. APCER$_{0.2}$ values are given in brackets.}
	\label{fig:det_fakefinger}
\end{figure}
For the following LOO experiments, we present the DET plots for all evaluated algorithms and additionally analyse the two best laser and two best SWIR methods as well as the fusion in more detail. To that end, we also list the corresponding APCEs.
Starting with the fakefinger group, Fig.~\ref{fig:det_fakefinger} shows the results for training on overlay attacks only. The best algorithms have slightly higher APCER$_{0.2}$ values than on the baseline partition (e.g.\ SWIR CNN VGG16: 4.35\%, +1\%) and our fusion achieves with 5.61\% only the fifth place (+3\%). On the other hand, SWIR CNN VGG19 improves by 2\% to 4.35\%. The number of APCEs for the best algorithms are presented in Table~\ref{tab:apces_fakefinger}. While the SWIR algorithms only struggle to detect orange playdoh fingers, the laser algorithms additionally have problems with dragonskin fingers.
Furthermore, we have three curves for the LRCN and both autoencoders that show considerable worse performance. While the LRCN should usually be able to detect that there is no blood movement, it suffers from the same type of attacks as the autoencoders~\cite{Kolberg-Autoencoder-FingerPAD-ARXIV-2020} and other laser CNNs: dragon skin as well as yellow and orange playdoh fingers are hardly classified as attack presentations. On the other hand, different playdoh colours are always correctly classified.
\begin{table}[t]
	\centering
	\setlength{\tabcolsep}{5.5pt}
	\caption{Number of APCEs at an APCER$_{0.2}$ on the \textit{Fakefinger} partition.}
	\label{tab:apces_fakefinger}
	\begin{tabular}{lccccc}
		\toprule
		\multirow{2}{*}{PAI} & \multicolumn{2}{c}{Laser} & \multicolumn{2}{c}{SWIR} & \multirow{1}{*}{Fusion} \\
		& ResNet & VGG16 & VGG16 & VGG19 & \\
		\midrule
		dragonskin & 30 & 41 & 3 & 6 & 5 \\
		ecoflex & 0 & 1 & 0 & 0 & 0 \\
		latex & 0 & 5 & 1 & 0 & 0 \\
		playdoh & 39 & 16 & 48 & 47 & 66 \\
		silly putty & 1 & 5 & 1 & 2 & 0 \\
		wax & 0 & 1 & 2 & 0 & 0 \\
		\midrule
		total & 70 & 70 & 55 & 55 & 71 \\
		\bottomrule
	\end{tabular}
\end{table}

\begin{figure}[t]
	\centering
	\includegraphics[width=0.87\linewidth]{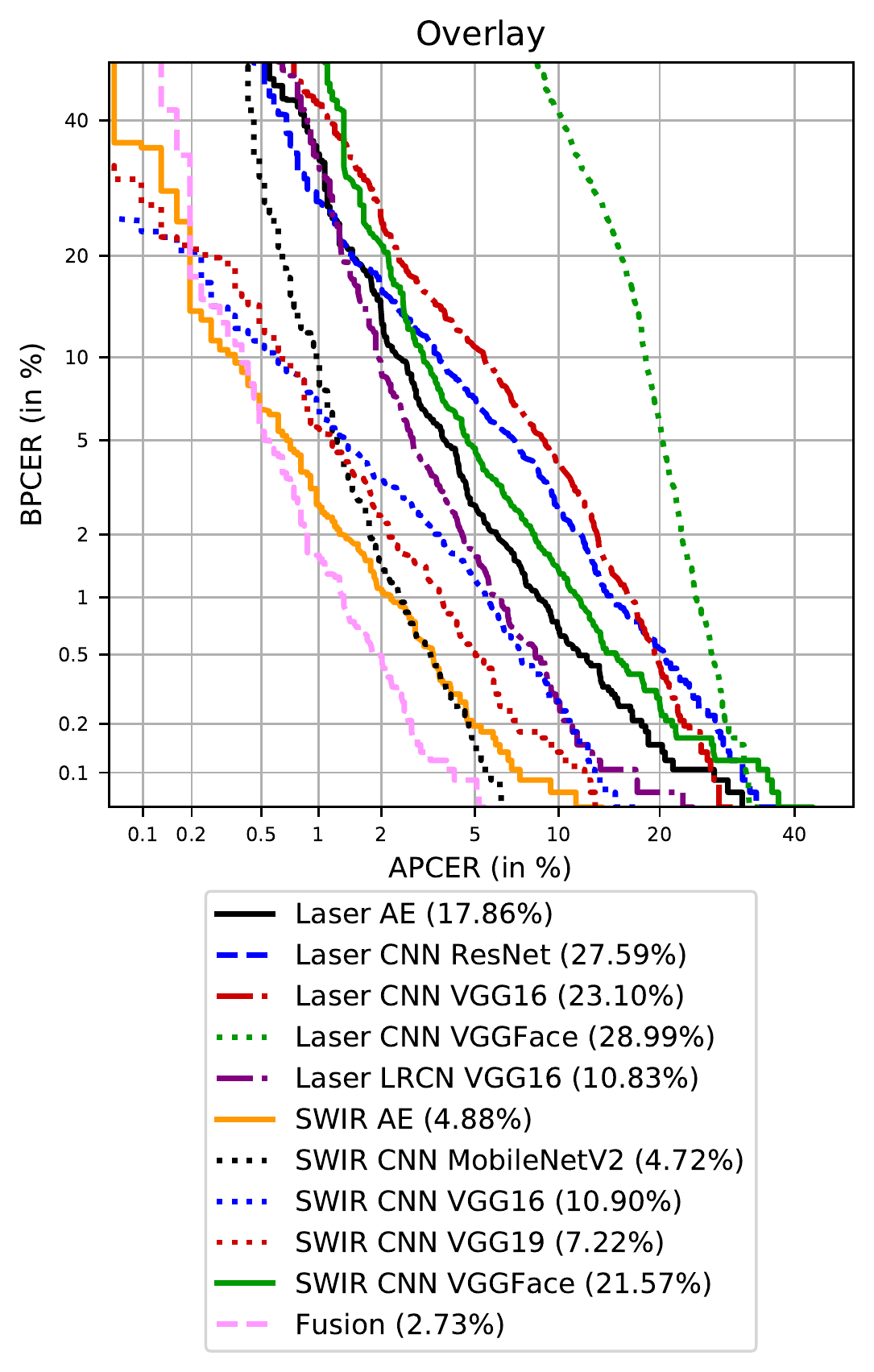}
	\caption{DET curves of the \emph{Overlay} group. APCER$_{0.2}$ values are given in brackets.}
	\label{fig:det_overlay}
\end{figure}
The DET plot for training only on fakefingers and leaving all overlays for testing is shown in Fig.~\ref{fig:det_overlay}. In this scenario, the autoencoders (especially SWIR) perform much better (3rd place) and the fusion achieves the best results, followed by the SWIR CNN MobileNetV2. However, the laser CNN VGGFace (part of the fusion) has the worst detection performance, hence fusing different algorithms would probably further improve the results. Detailed information on misclassified PAs are given in Table~\ref{tab:apces_overlay}. The biggest challenge to all algorithms are transparent silicone PAIs when not seen in training.
\begin{table}[t]
	\centering
	\setlength{\tabcolsep}{6pt}
	\caption{Number of APCEs at an APCER$_{0.2}$ on the \textit{Overlay} partition.}
	\label{tab:apces_overlay}
	\begin{tabular}{lccccc}
		\toprule
		\multirow{2}{*}{PAI} & \multicolumn{2}{c}{Laser} & \multicolumn{2}{c}{SWIR} & Fusion \\
		& AE & LRCN & AE & MobileNetV2 & \\
		\midrule
		\multicolumn{6}{l}{\textit{opaque}} \\
		bandage & 2 & 5 & 1 & 2 & 0 \\
		dragonskin & 16 & 4 & 0 & 0 & 0 \\
		gelatin & 95 & 20 & 0 & 0 & 0 \\
		silicone & 10 & 70 & 0 & 2 & 1 \\
		\midrule
		\multicolumn{6}{l}{\textit{transparent}} \\
		dragonskin & 58 & 39 & 21 & 10 & 5 \\ 
		gelatin & 44 & 33 & 13 & 47 & 21 \\
		glue & 22 & 18 & 10 & 3 & 2 \\
		latex & 20 & 8 & 0 & 0 & 0 \\
		silicone & 152 & 113 & 97 & 74 & 53 \\
		wax & 2 & 0 & 0 & 0 & 0 \\
		\midrule
		\multicolumn{6}{l}{\textit{semi}} \\
		dragonskin & 42 & 15 & 0 & 5 & 2 \\
		ecoflex & 21 & 5 & 2 & 2 & 0 \\
		glue & 60 & 2 & 6 & 0 & 0 \\
		silicone & 4 & 1 & 0 &  0 & 0 \\
		\midrule
		total & 549 & 333 & 150 & 145 & 84 \\
		\bottomrule
	\end{tabular}
\end{table}

\begin{figure}[t]
	\centering
	\includegraphics[width=0.87\linewidth]{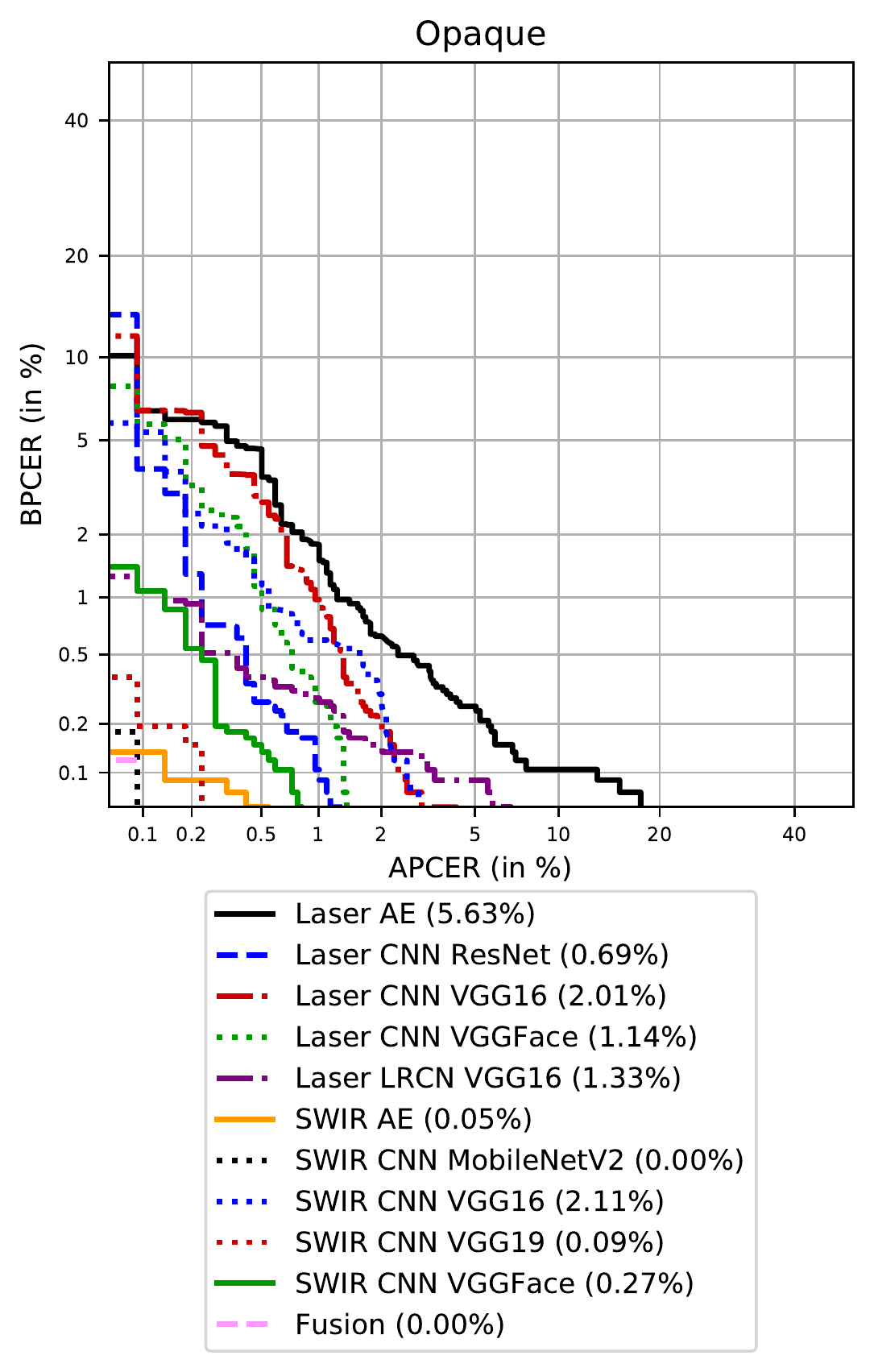}
	\caption{DET curves of the \emph{Opaque} group. APCER$_{0.2}$ values are given in brackets.}
	\label{fig:det_opaque}
\end{figure}
Far better results are obtained for the opaque group (Fig.~\ref{fig:det_opaque}) since the training includes even more transparent overlay PAIs. The fusion as well as the SWIR CNN MobileNetV2 are able to correctly classify all PA samples for a BPCER of 0.2\%. The worst algorithm reports an APCER$_{0.2}$ of 5.63\% while all others are at an APCER$_{0.2}$ around or below 2\%. Thus, we can conclude so far that opaque overlay attacks are no threat to the utilised capture device. Particular numbers of APCEs for this group are presented in Table~\ref{tab:apces_opaque}, showing a slight weakness for the laser algorithms when facing ecoflex PAIs. However, in contrast to the opaque results within the overlay partition (Table~\ref{tab:apces_overlay}), we see a big performance improvement, especially for the laser algorithms, when leaving only opaque overlays for testing.
\begin{table}[t]
	\centering
	\setlength{\tabcolsep}{5.5pt}
	\caption{Number of APCEs at an APCER$_{0.2}$ on the \textit{Opaque} partition.}
	\label{tab:apces_opaque}
	\begin{tabular}{lccccc}
		\toprule
		\multirow{2}{*}{PAI} & \multicolumn{2}{c}{Laser} & \multicolumn{2}{c}{SWIR} & Fusion \\
		& ResNet & VGGFace & AE & MobileNetV2 & \\
		\midrule
		bandage & 0 & 0 & 1 & 0 & 0 \\
		ecoflex & 7 & 21 & 0 & 0 & 0 \\
		gelatin & 5 & 1 & 0 & 0 & 0 \\
		silicone & 3 & 3 & 0 & 0 & 0 \\ 
		\midrule
		total & 15 & 25 & 1 & 0 & 0 \\
		\bottomrule
	\end{tabular}
\end{table}

\begin{figure}[t]
	\centering
	\includegraphics[width=0.87\linewidth]{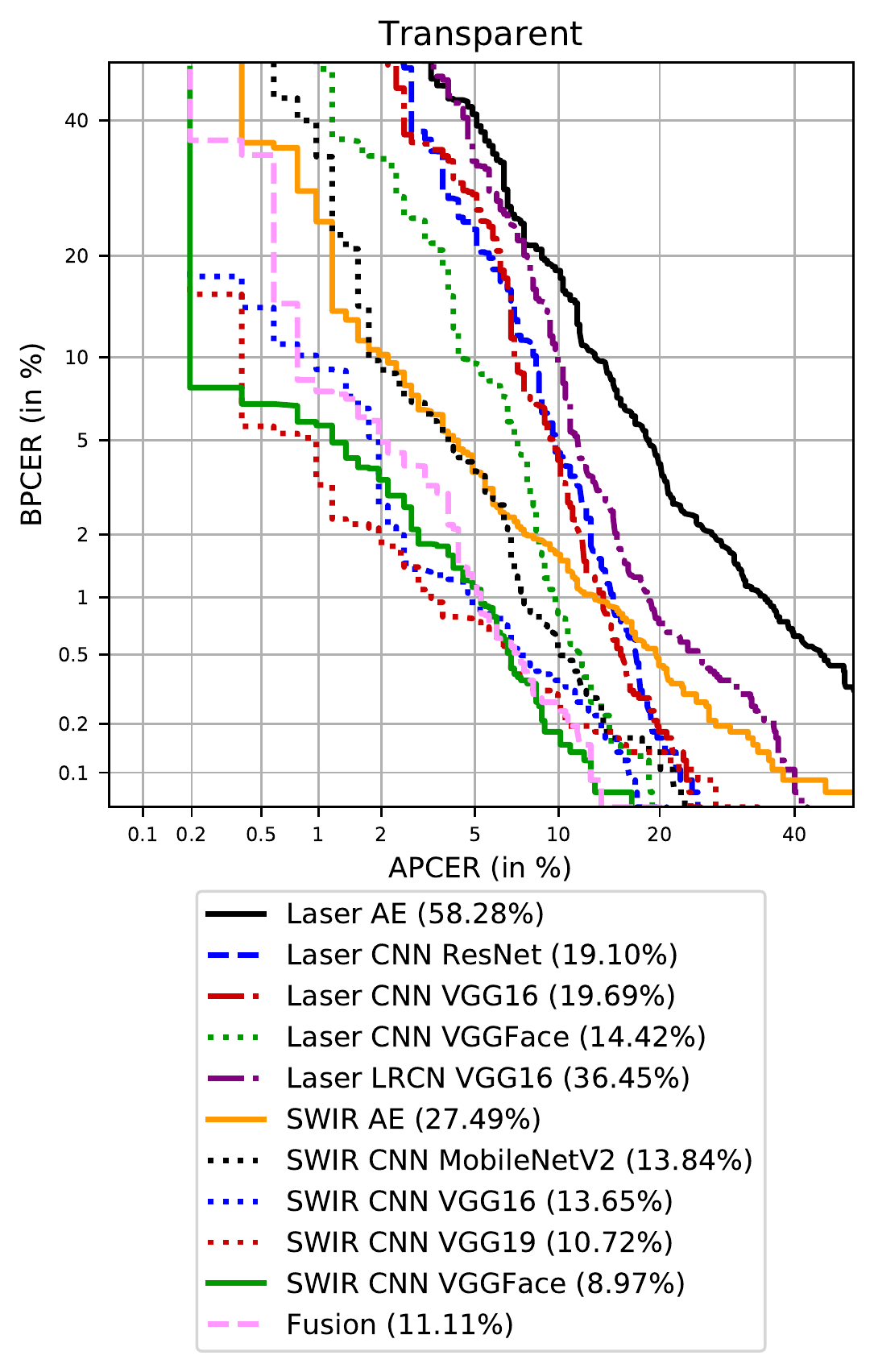}
	\caption{DET curves of the \emph{Transparent} group. APCER$_{0.2}$ values are given in brackets.}
	\label{fig:det_transparent}
\end{figure}
The highest error rates on the other hand are shown in Fig.~\ref{fig:det_transparent} for the transparent group. Hence, the most challenging PAD task is to detect thin and transparent overlay attacks when they are unknown during the training. The best performing network is the VGGFace for both laser and SWIR data, which also beats the fusion for SWIR. As expected, the LRCN still detects blood movement behind the overlays as in \cite{Kolberg-LSTM-FingerPAD-IJCB-2020}, which leads to higher classification errors. Table~\ref{tab:apces_tranparent} lists detailed numbers of APCEs proving that transparent silicone PAIs are accountable for the major part of misclassified samples. Further occurring errors are not as significant since the fusion detects most of them. In this context, we see a big improvement of the fusion in detecting gelatin and dragonskin PAIs in comparison to the overlay partition results in Table~\ref{tab:apces_overlay}.
\begin{table}[t]
	\centering
	\setlength{\tabcolsep}{4.5pt}
	\caption{Number of APCEs at an APCER$_{0.2}$ on the \textit{Transparent} partition.}
	\label{tab:apces_tranparent}
	\begin{tabular}{lccccc}
		\toprule
		\multirow{2}{*}{PAI} & \multicolumn{2}{c}{Laser} & \multicolumn{2}{c}{SWIR} & Fusion \\
		& ResNet & VGGFace & VGG19 & VGGFace & \\
		\midrule
		dragonskin & 15 & 6 & 17 & 2 & 0 \\ 
		gelatin & 2 & 5 & 1 & 0 & 4 \\
		glue & 7 & 9 & 1 & 1 & 1 \\
		latex & 0 & 0 & 0 & 1 & 0 \\
		silicone & 74 & 54 & 36 & 41 & 50 \\
		\midrule
		total & 98 & 74 & 55 & 46 & 55 \\
		\bottomrule
	\end{tabular}
\end{table}

\begin{figure}[t]
	\centering
	\includegraphics[width=0.87\linewidth]{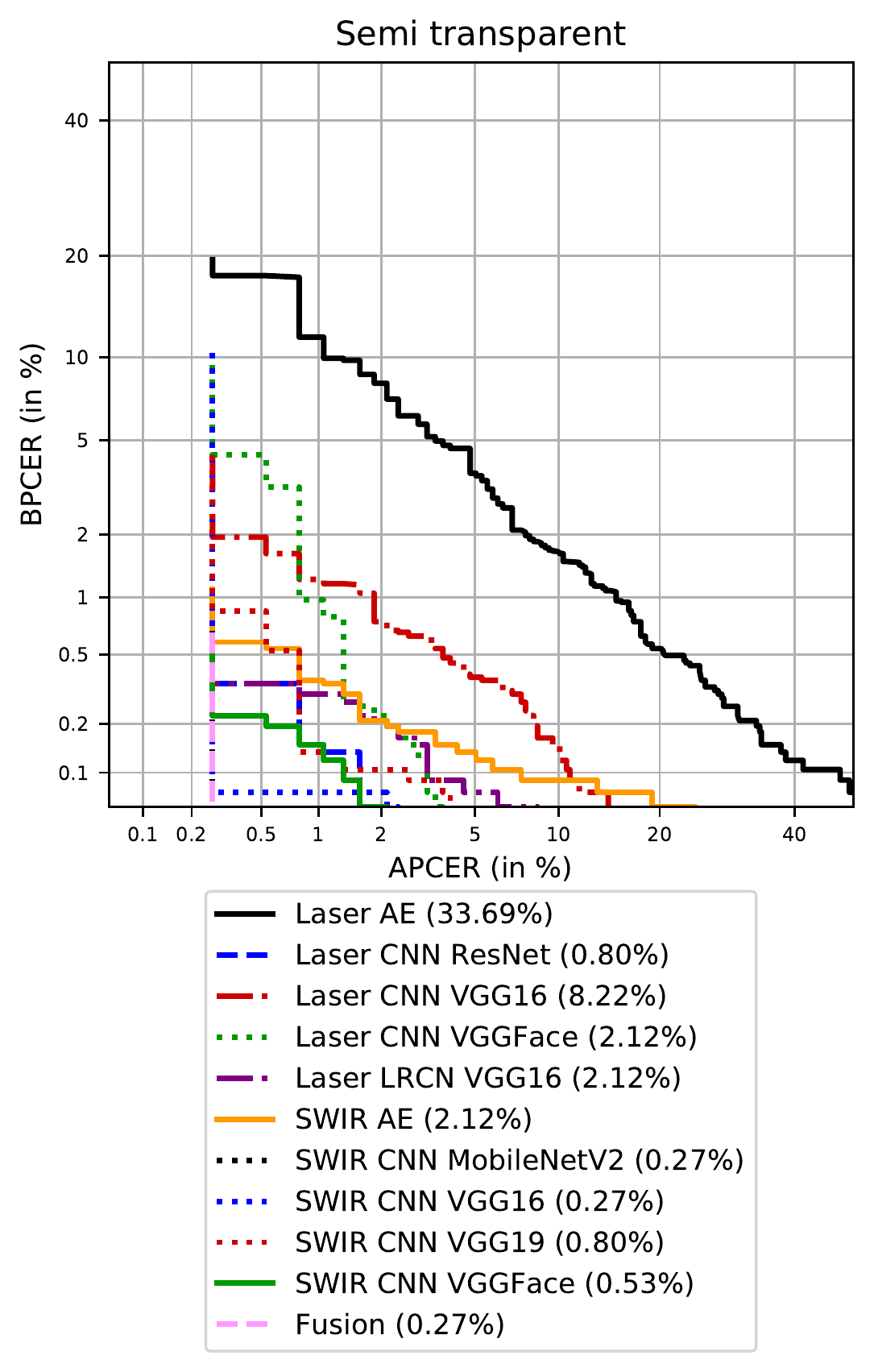}
	\caption{DET curves of the \emph{Semi transparent} group. APCER$_{0.2}$ values are given in brackets.}
	\label{fig:det_semi}
\end{figure}
Aligning with the previous experiments, the semi transparent results settle between the opaque and transparent ones as depicted in Fig.~\ref{fig:det_semi}. As the PAD performance is closer to the opaque one, the fusion is among the best-performing algorithms while most APCER$_{0.2}$ values are below 2.2\%. APCEs stem from dragonskin, ecoflex, glue, or silicone PAIs as depicted in Table~\ref{tab:apces_semi}. Moreover, also the semi transparent results improve towards the overlay results of Table~\ref{tab:apces_overlay}.
\begin{table}[t]
	\centering
	\setlength{\tabcolsep}{3pt}
	\caption{Number of APCEs at an APCER$_{0.2}$ for \textit{Semi transparent} partition.}
	\label{tab:apces_semi}
	\begin{tabular}{lccccc}
		\toprule
		\multirow{2}{*}{PAI} & \multicolumn{2}{c}{Laser} & \multicolumn{2}{c}{SWIR} & Fusion \\
		& ResNet & VGGFace & MobileNetV2 & VGG16 & \\
		\midrule
		dragonskin & 0 & 1 & 0 & 0 & 0 \\
		ecoflex & 0 & 0 & 1 & 0 \\
		glue & 0 & 2 & 0 & 0 & 0 \\
		silicone & 3 & 5 & 1 & 0 & 1 \\
		\midrule
		total & 3 & 8 & 1 & 1 & 1 \\
		\bottomrule
	\end{tabular}
\end{table}

Until now we have seen that there are no big differences whether the PAD algorithms are trained on all PAI species or only on fakefingers or overlays, respectively. However, detecting unknown opaque overlays is much easier than unknown transparent ones. Hence, the similar performance to the baseline system are due to the specific shares of samples from the different PAI groups. Additionally, for these LOO experiments, the pre-fixed fusion is among the best algorithms. Thus, while single algorithms report a more unsteady PAD performance, the fusion generalises much better across unknown scenarios. Another lessons learned is that the LRCN fails to detect fakefinger PAIs without blood movement when it is trained only on overlay PAIs. Hence, the performance of the LRCN highly depends on the training set and does not really generalise.

\subsection{Material LOO Groups}
\begin{figure}[t]
	\centering
	\includegraphics[width=0.87\linewidth]{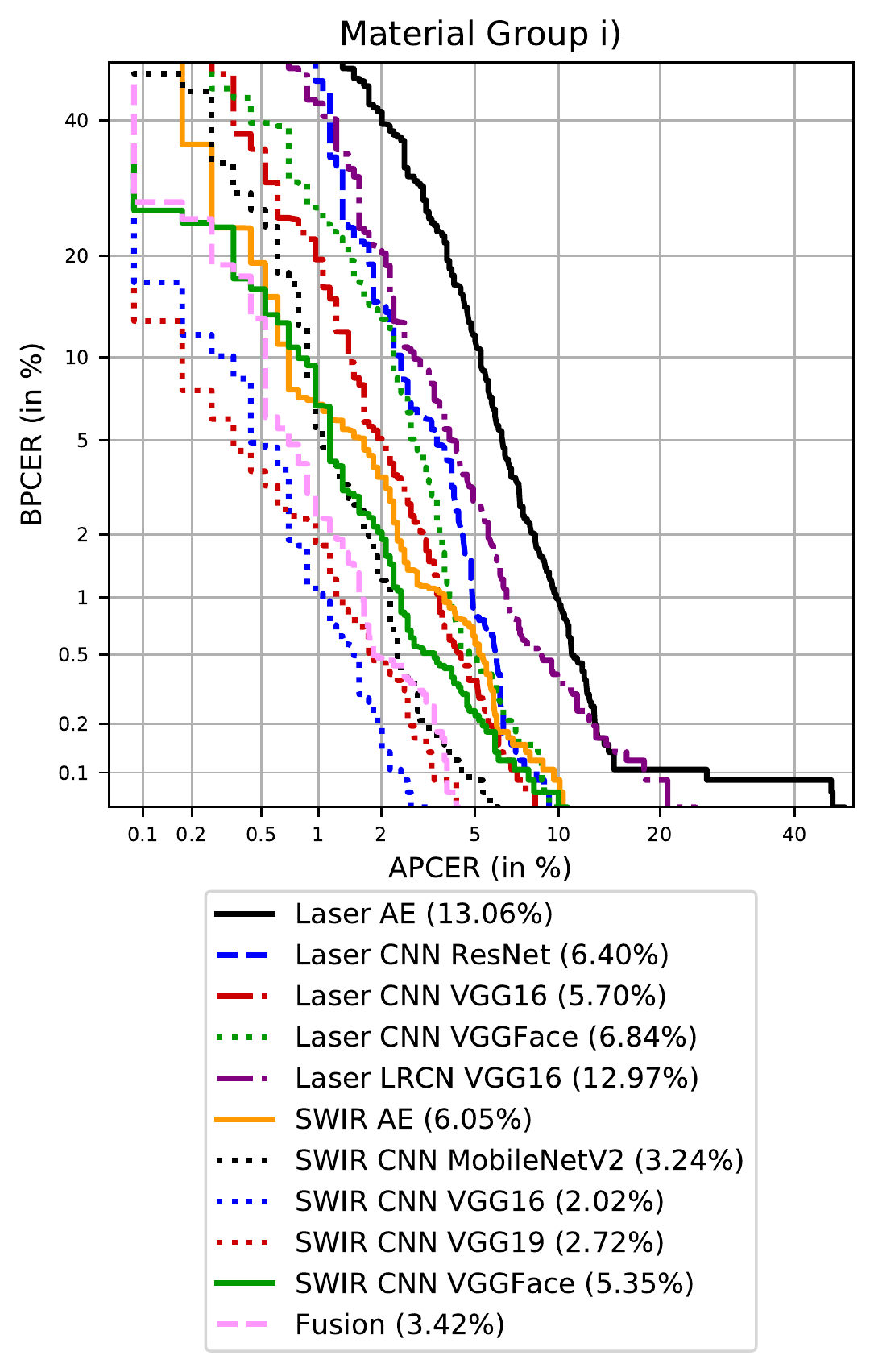}
	\caption{DET curves of the material group i). APCER$_{0.2}$ values are given in brackets.}
	\label{fig:det_g1}
\end{figure}
After focussing on visual properties, the subsequent part evaluates the four material groups that are subsequently left out from training. The DET plot for material group i), containing all silicone PAIs, is shown in Fig.~\ref{fig:det_g1}. The best results are obtained by SWIR CNNs VGG16 and VGG19, followed by SWIR CNN MobileNetV2 and our fusion (all APCERs$_{0.2}$ below 3.5\%). When having a look at the differences for laser and SWIR algorithms regarding APCEs in Table~\ref{tab:apces_g1} we see that the majority of misclassifications occur for transparent silicone PAIs. Moreover, the best-performing laser CNNs ResNet and VGG16 have approximately twice as much errors for transparent PAIs as the SWIR CNNs. Hence, the fusion performs also slightly worse than some SWIR algorithms on its own.
\begin{table}[t]
	\centering
	\setlength{\tabcolsep}{5pt}
	\caption{Number of APCEs at an APCER$_{0.2}$ for material group i) partition.}
	\label{tab:apces_g1}
	\begin{tabular}{lccccc}
		\toprule
		\multirow{2}{*}{PAI} & \multicolumn{2}{c}{Laser} & \multicolumn{2}{c}{SWIR} & Fusion \\
		& ResNet & VGG16 & VGG16 & VGG19 & \\
		\midrule
		opaque & 4 & 6 & 1 & 1 & 0 \\
		transparent & 68 & 55 & 22 & 30 & 39 \\
		semi & 1 & 4 & 0 & 0 & 0 \\
		\midrule
		total & 73 & 65 & 23 & 31 & 39 \\
		\bottomrule
	\end{tabular}
\end{table}

\begin{figure}[t]
	\centering
	\includegraphics[width=0.87\linewidth]{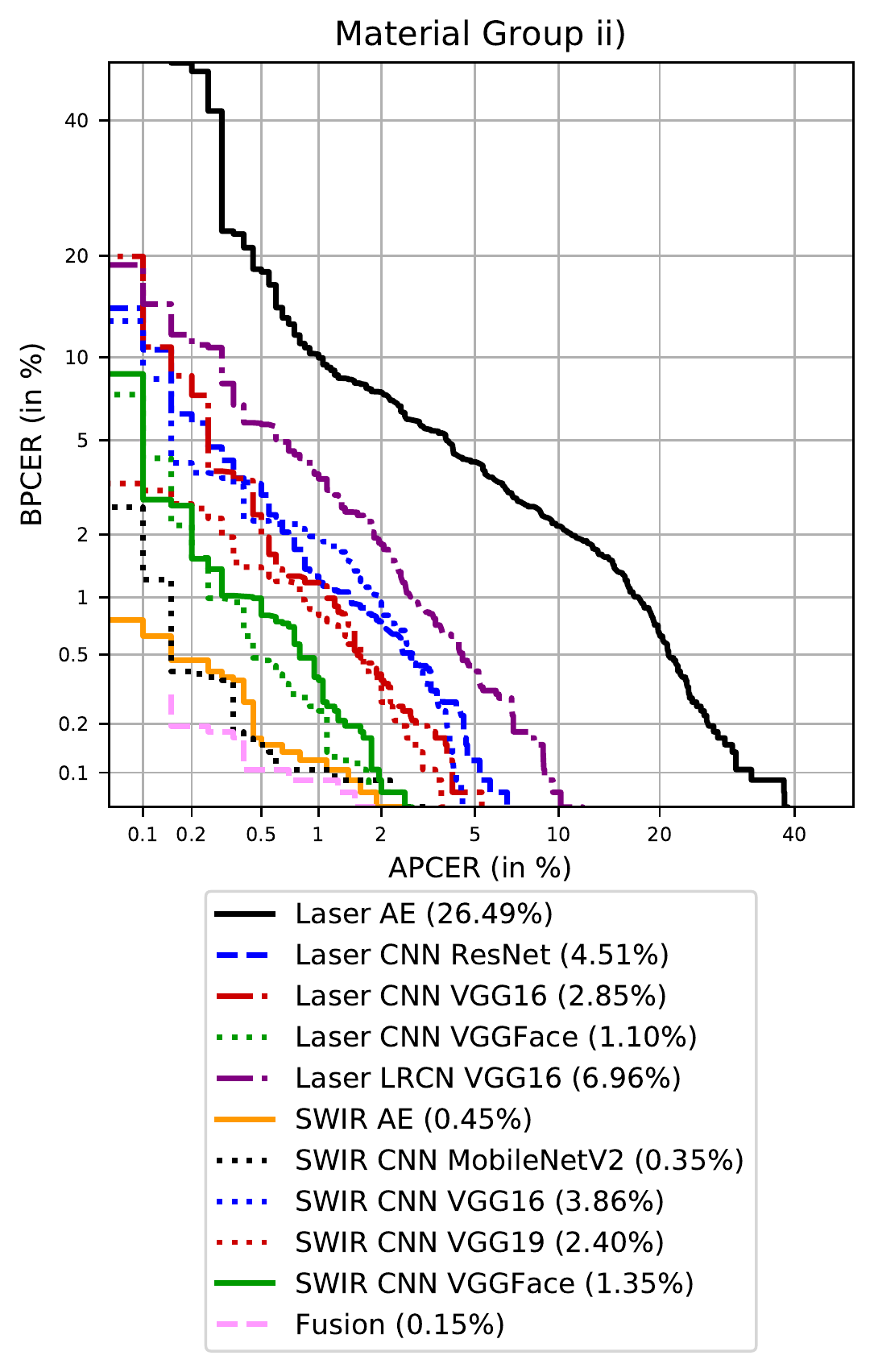}
	\caption{DET curves of the material group ii). APCER$_{0.2}$ values are given in brackets.}
	\label{fig:det_g2}
\end{figure}
The second material group left out dragonskin and ecoflex for training since their appearance and fabrication are very similar. The resulting DET is plotted in Fig.~\ref{fig:det_g2}. Compared to the first material group (Fig.~\ref{fig:det_g1}), nearly all DET curves shifted to the left, reflecting better results. The fusion misclassifies only three PAIs (0.15\%), followed by SWIR CNN MobileNetV2, SWIR AE, and laser CNN VGGFace. Table~\ref{tab:apces_g2} reveals that especially for laser algorithms more APCEs stem from dragonskin PAIs. This is due to the fact that using ecoflex, no fully transparent overlays are in the dataset (see Table~\ref{tab:pais}). Since, the laser penetrates thin and transparent overlays, these samples appear very similar to bona fide ones and can be misclassified.
\begin{table}[t]
	\centering
	\setlength{\tabcolsep}{4pt}
	\caption{Number of APCEs at an APCER$_{0.2}$ for material group ii) partition.}
	\label{tab:apces_g2}
	\begin{tabular}{lccccc}
		\toprule
		\multirow{2}{*}{PAI} & \multicolumn{2}{c}{Laser} & \multicolumn{2}{c}{SWIR} & Fusion \\
		& VGG16 & VGGFace & AE & MobileNetV2 & \\
		\midrule
		dragonskin & 41 & 19 & 9 & 4 & 1 \\
		ecoflex & 16 & 2 & 0 & 3 & 2 \\
		\midrule
		total & 57 & 21 & 9 & 7 & 3 \\
		\bottomrule
	\end{tabular}
\end{table}

\begin{figure}[t]
	\centering
	\includegraphics[width=0.87\linewidth]{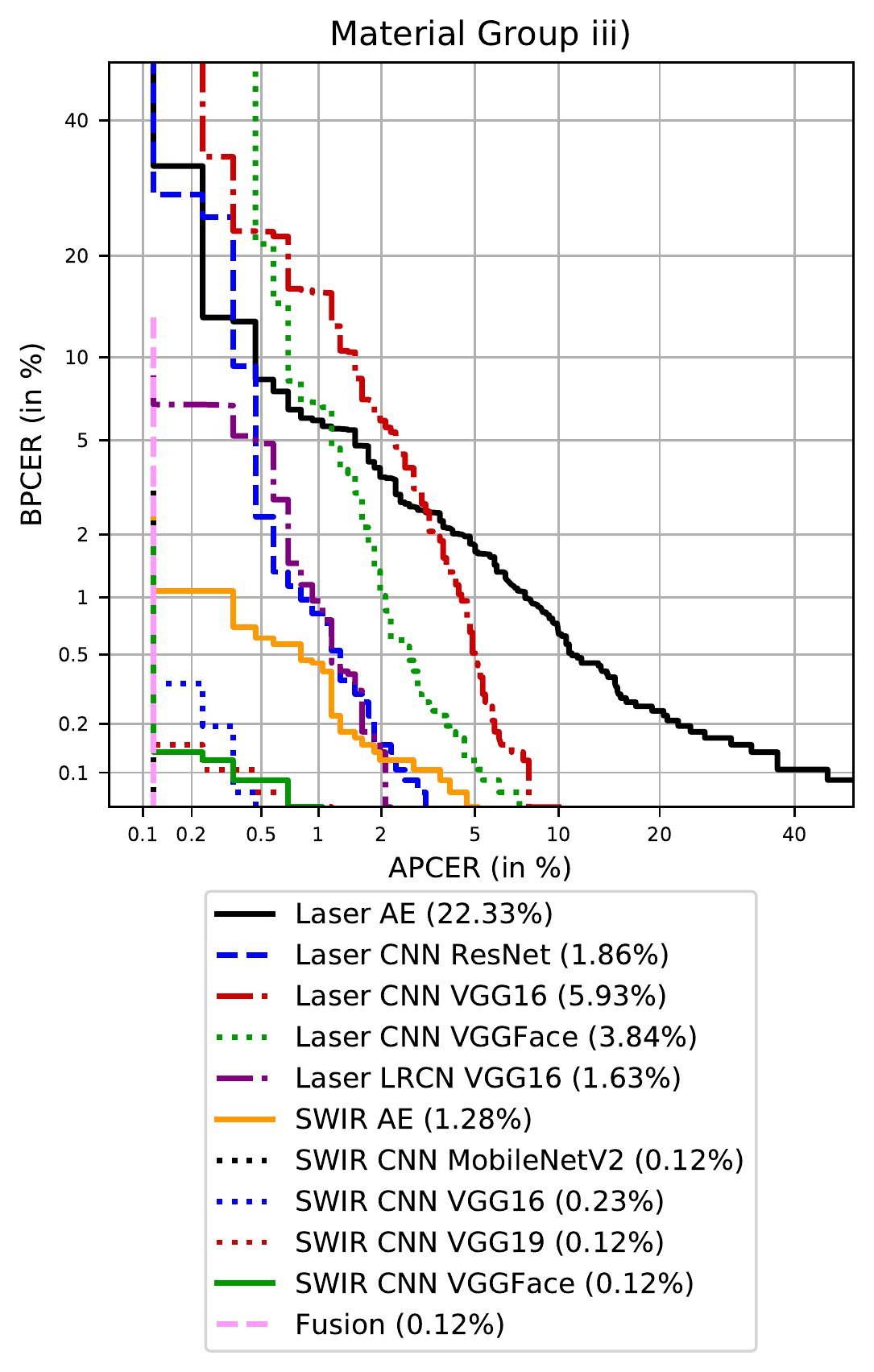}
	\caption{DET curves of the material group iii). APCER$_{0.2}$ values are given in brackets.}
	\label{fig:det_g3}
	\vspace{-1ex}
\end{figure}
The third group combines PAIs made from gelatin, glue, latex, printouts, and wax (Fig.~\ref{fig:det_g3}). This group achieves the best results from all material groups. Especially the SWIR CNNs and the fusion correctly classify all PAI samples but one or two for the given BPCER = 0.2\%. On the other hand, LRCN VGG16 (1.63\%) and CNN ResNet (1.86\%) achieve the best performance on the laser data. Table~\ref{tab:apces_g3} shows that most APCEs occur for glue while gelatin at least troubles the laser CNN to the same extend. In general these left out materials are much easier to detect than the ones from the other groups.
\begin{table}[t]
	\centering
	\setlength{\tabcolsep}{5pt}
	\caption{Number of APCEs at an APCER$_{0.2}$ for material group iii) partition.}
	\label{tab:apces_g3}
	\begin{tabular}{lccccc}
		\toprule
		\multirow{2}{*}{PAI} & \multicolumn{2}{c}{Laser} & \multicolumn{2}{c}{SWIR} & Fusion \\
		& ResNet & LRCN & MobileNetV2 & VGG19 & \\
		\midrule
		gelatin & 7 & 2 & 0 & 0 & 0 \\
		glue & 8 & 11 & 1 & 1 & 1 \\
		latex & 1 & 1 & 0 & 0 & 0 \\
		\midrule
		total & 16 & 14 & 1 & 1 & 1 \\
		\bottomrule
	\end{tabular}
\end{table}

\begin{figure}[t]
	\centering
	\includegraphics[width=0.87\linewidth]{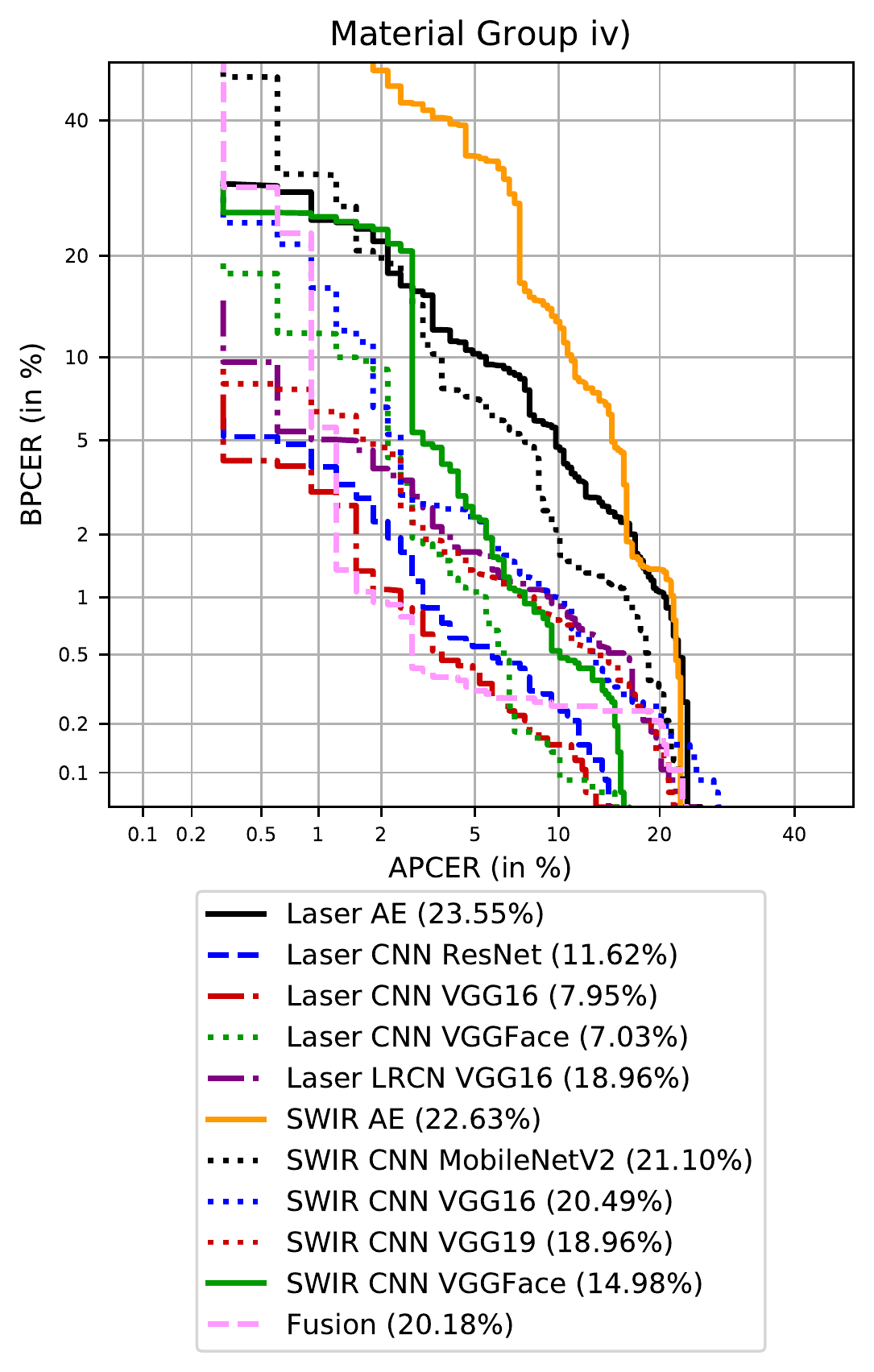}
	\caption{DET curves of the material group iv). APCER$_{0.2}$ values are given in brackets.}
	\label{fig:det_g4}
	\vspace{-1em}
\end{figure}
The last material group includes samples from 3D printed, dental material, playdoh, and silly putty PAIs. The DET plot in Fig.~\ref{fig:det_g4} shows the worst performance from all material groups. This is related to the small test set such that one error effects the total APCER much more. Furthermore, this is the only experiment where SWIR results (average 19.6\%), and the fusion, are significantly worse than the laser results (average 13.8\%). As listed in Table~\ref{tab:apces_g4}, the reason are playdoh PAIs. While SWIR CNNs misclassify 57 (VGG19) and 49 (VGGFace) playdoh samples, their laser counterparts misclassify only 14 (VGG16) and 8 (VGGFace). On the other hand, the laser CNNs fail to detect further samples from silly putty. As previous studies already showed \cite{GomezBarrero-MSSWIRCNN-CRC-2020, Tolosana-SWIR-PAD-CNNs-TIFS-2020}, orange playdoh reflects the four SWIR wavelengths nearly identical as bona fide skin.
\begin{table}[t]
	\centering
	\setlength{\tabcolsep}{3pt}
	\caption{Number of APCEs at an APCER$_{0.2}$ for material group iv) partition.}
	\label{tab:apces_g4}
	\begin{tabular}{lccccc}
		\toprule
		\multirow{2}{*}{PAI} & \multicolumn{2}{c}{Laser} & \multicolumn{2}{c}{SWIR} & Fusion \\
		& VGG16 & VGGFace & VGG19 & VGGFace & \\
		\midrule
		dental material & 1 & 0 & 0 & 0 & 0 \\
		playdoh & 14 & 8 & 57 & 49 & 66 \\
		silly putty & 11 & 15 & 5 & 0 & 0 \\
		\midrule
		total & 26 & 23 & 62 & 49 & 66 \\
		\bottomrule
	\end{tabular}
	\vspace{-1ex}
\end{table}

All in all, analysing the material groups confirms observations already made for the previous LOO experiments: \textit{i)} transparent PAIs are more challenging to detect, \textit{ii)} SWIR algorithms perform generally better than laser algorithms (except for playdoh), \textit{iii)} the pre-defined fusion generalises quiet well and achieves in most cases the lowest error rates. The most challenging materials per group are silicone, dragonskin, glue, and playdoh. However, this additionally depends on more specific properties as e.g.\ colour and thickness.

\section{Conclusions}
\label{sec:conclusions}
In this work, we benchmarked the generalisation capabilities of fingerprint PAD methods on a database captured in the short wave infrared domain that includes more than 24,000 samples comprising 45 PAI species. To that end, we grouped the PAI species in five clusters that are relevant for this capture device: fakefinger, all overlays, opaque overlays, transparent overlays, and semi transparent overlays. Additionally, four material-specific groups are selected.
For each experiment we left one group exclusively for testing and trained only on all other groups. Thus, we were able to compare the PAD performance on unknown attack species across ten single PAD approaches and one combined fusion. 
As a result, we observed that opaque overlay attacks are no challenge to this PAD approach (APCER$_{0.2}=0$\% in the best case) but the risk increases for transparent overlays (APCER$_{0.2}\approx9$\% in the best case). In the fakefinger group, mostly orange playdoh fingers trouble the classifier since their appearance resembles very much the bona fide samples in the SWIR domain. As soon as these PAIs are not included in the training set, the classifier does not learn the subtle differences.
Finally, the pre-defined fusion of one laser and one SWIR algorithm generalises quite well even if one fusion partner performs much worse than on the baseline partition. While this fusion performs best on the baseline partition (APCER$_{0.2}$ = 1.81\%), it further achieves the best performance across several LOO experiments. For the other cases (fakefinger, transparent, \nth{1} material group) the fusion reports about a 1.5\% higher APCER$_{0.2}$ than the best algorithm. However, these results are still among the best (\nth{5} for fakefinger, \nth{2} for transparent, and \nth{4} for \nth{1} material group). The only outlier is the \nth{4} material group (\nth{7} with +13\% APCER$_{0.2}$ compared to the best algorithm).

\section{Acknowledgments}
This research is based upon work supported in part by the Office of the Director of National Intelligence (ODNI), Intelligence Advanced Research Projects Activity (IARPA) under contract number 2017-17020200005. The views and conclusions contained herein are those of the authors and should not be interpreted as necessarily representing the official policies, either expressed or implied, of ODNI, IARPA, or the U.S.\ Government. The U.S.\ Government is authorized to reproduce and distribute reprints for governmental purposes notwithstanding any copyright annotation therein.
Furthermore, we would like to thank our colleagues from Information Science Institute at University of Southern California for the data collection effort.
This research work has been funded by the German Federal Ministry of Education and Research and the Hessian Ministry of Higher Education, Research, Science and the Arts within their joint support of the National Research Center for Applied Cybersecurity ATHENE.





\bibliographystyle{IEEEtran}
\bibliography{loo}

\begin{thebibliography}{10}
\providecommand{\url}[1]{#1}
\csname url@samestyle\endcsname
\providecommand{\newblock}{\relax}
\providecommand{\bibinfo}[2]{#2}
\providecommand{\BIBentrySTDinterwordspacing}{\spaceskip=0pt\relax}
\providecommand{\BIBentryALTinterwordstretchfactor}{4}
\providecommand{\BIBentryALTinterwordspacing}{\spaceskip=\fontdimen2\font plus
\BIBentryALTinterwordstretchfactor\fontdimen3\font minus
  \fontdimen4\font\relax}
\providecommand{\BIBforeignlanguage}[2]{{%
\expandafter\ifx\csname l@#1\endcsname\relax
\typeout{** WARNING: IEEEtran.bst: No hyphenation pattern has been}%
\typeout{** loaded for the language `#1'. Using the pattern for}%
\typeout{** the default language instead.}%
\else
\language=\csname l@#1\endcsname
\fi
#2}}
\providecommand{\BIBdecl}{\relax}
\BIBdecl

\bibitem{Jain-BiometricRecognition-Nature-2007}
A.~K. Jain, ``Technology: Biometric recognition,'' \emph{Nature}, vol. 449, no.
  7158, p.~38, 2007.

\bibitem{ISO-IEC-30107-1-PAD-Framework-160115}
{ISO/IEC JTC1 SC37 Biometrics}, \emph{{ISO/IEC} 30107-1. Information Technology
  - Biometric presentation attack detection - Part 1: Framework}, 2016.

\bibitem{Biggio-EvaluationOfPAs-IETBiometrics-2012}
B.~Biggio, Z.~Akhtar, G.~Fumera, G.~L. Marcialis, and F.~Roli, ``Security
  evaluation of biometric authentication systems under real spoofing attacks,''
  \emph{IET Biometrics}, vol.~1, no.~1, pp. 11--24, 2012.

\bibitem{Husseis-SurveyPA+PAD-ICCST-2019}
A.~Husseis, J.~Liu-Jimenez, I.~Goicoechea-Telleria, and R.~Sanchez-Reillo, ``A
  survey in presentation attack and presentation attack detection,'' in
  \emph{Proc. Intl. Carnahan Conference on Security Technology
  ({ICCST})}.\hskip 1em plus 0.5em minus 0.4em\relax IEEE, 2019, pp. 1--13.

\bibitem{Marcel-HandbookPAD-ACVPR-2019}
S.~Marcel, M.~S. Nixon, J.~Fierrez, and N.~Evans, \emph{Handbook of Biometric
  Anti-Spoofing: Presentation Attack Detection}.\hskip 1em plus 0.5em minus
  0.4em\relax Springer, 2019.

\bibitem{Sousedik-PAD-Survey-IET-BMT-2014}
C.~Sousedik and C.~Busch, ``Presentation attack detection methods for
  fingerprint recognition systems: A survey,'' \emph{IET Biometrics}, vol.~3,
  no.~1, pp. 1--15, 2014.

\bibitem{Marasco-PAD-SurveyFingerprint-CSUR-2014}
E.~Marasco and A.~Ross, ``A survey on antispoofing schemes for fingerprint
  recognition systems,'' \emph{ACM Computing Surveys (CSUR)}, vol.~47, no.~2,
  pp. 1--36, 2014.

\bibitem{Kanich-FingerprintPAIs-IWBF-2018}
O.~Kanich, M.~Drahansky, and M.~M{\'e}zl, ``Use of creative materials for
  fingerprint spoofs,'' in \emph{Proc. Intl. Workshop on Biometrics and
  Forensics (IWBF)}, 2018.

\bibitem{Tan-EffectUnknownPAIs-FingerPAD-WIFS-2010}
B.~Tan, A.~Lewicke, D.~Yambay, and S.~Schuckers, ``The effect of environmental
  conditions and novel spoofing methods on fingerprint anti-spoofing
  algorithms,'' in \emph{Proc. Intl. Workshop on Information Forensics and
  Security ({WIFS)}}.\hskip 1em plus 0.5em minus 0.4em\relax IEEE, 2010, pp.
  1--6.

\bibitem{Marasco-RobustnessUnknownPAI-FingerPAD-BIOSIGNALS-2011}
E.~Marasco and C.~Sansone, ``On the robustness of fingerprint liveness
  detection algorithms against new materials used for spoofing,'' in
  \emph{BIOSIGNALS}, vol.~8, 2011, pp. 553--555.

\bibitem{Rattani-GeneralisationFingerprintPAD-TIFS-2015}
A.~Rattani, W.~J. Scheirer, and A.~Ross, ``Open set fingerprint spoof detection
  across novel fabrication materials,'' \emph{IEEE Trans. on Information
  Forensics and Security ({TIFS})}, vol.~10, no.~11, pp. 2447--2460, 2015.

\bibitem{Ding-OneClassFingerPAD-WIFS-2016}
Y.~Ding and A.~Ross, ``An ensemble of one-class {SVMs} for fingerprint spoof
  detection across different fabrication materials,'' in \emph{Proc. Intl.
  Workshop on Information Forensics and Security (WIFS)}.\hskip 1em plus 0.5em
  minus 0.4em\relax IEEE, 2016, pp. 1--6.

\bibitem{Nogueira-CNNFingerprintPAD-TIFS-2016}
R.~F. Nogueira, R.~{de Alencar Lotufo}, and R.~C. Machado, ``Fingerprint
  liveness detection using convolutional neural networks,'' \emph{IEEE Trans.
  on Information Forensics and Security ({TIFS})}, vol.~11, no.~6, pp.
  1206--1213, 2016.

\bibitem{Engelsma-OneClassFingerPAD-ICB-2019}
J.~J. Engelsma and A.~K. Jain, ``Generalizing fingerprint spoof detector:
  Learning a one-class classifier,'' in \emph{Proc. Intl. Conf. on Biometrics
  (ICB)}.\hskip 1em plus 0.5em minus 0.4em\relax IEEE, 2019, pp. 1--8.

\bibitem{Mirzaalian-LSCI-FingerPAD-2019}
H.~Mirzaalian, M.~Hussein, and W.~{Abd-Almageed}, ``On the effectiveness of
  laser speckle contrast imaging and deep neural networks for detecting known
  and unknown fingerprint presentation attacks,'' in \emph{Proc. Intl. Conf. on
  Biometrics (ICB)}, 2019, pp. 1--8.

\bibitem{Chugh-PADGeneralisation-ICB-2019}
T.~Chugh and A.~K. Jain, ``Fingerprint presentation attack detection:
  Generalization and efficiency,'' in \emph{Intl. Conf. on Biometrics
  (ICB)}.\hskip 1em plus 0.5em minus 0.4em\relax IEEE, 2019, pp. 1--8.

\bibitem{Chugh-FingerprintMaterialGenerator-TIFS-2020}
T.~Chugh and A.~Jain, ``Fingerprint spoof generalization,'' \emph{IEEE Trans.
  on Information Forensics and Security (TIFS)}, 2020.

\bibitem{GonzalezSoler-FingerPADFeatEncoding-TBIOM-2019}
L.~J. Gonz{\'a}lez-Soler, M.~Gomez-Barrero, L.~Chang, A.~Perez-Suarez,
  J.~Hernandez-Palancar, and C.~Busch, ``Fingerprint presentation attack
  detection based on local features encoding for unknown attacks,'' \emph{arXiv
  preprint https://arxiv.org/abs/1908.10163}, 2019.

\bibitem{Grosz-CrossSensorFingerprintPAD-Arxiv-2020}
S.~A. Grosz, T.~Chugh, and A.~K. Jain, ``Fingerprint presentation attack
  detection: A sensor and material agnostic approach,'' \emph{arXiv preprint
  arXiv:2004.02941}, 2020.

\bibitem{Kolberg-Autoencoder-FingerPAD-ARXIV-2020}
J.~Kolberg, M.~Grimmer, M.~Gomez-Barrero, and C.~Busch, ``Anomaly detection
  with convolutional autoencoders for fingerprint presentation attack
  detection,'' \emph{Arxiv preprint}, 2020.

\bibitem{Singh-SurveyUnknownFingerprintPAD-Arxiv-2020}
J.~M. Singh, A.~Madhun, G.~Li, and R.~Ramachandra, ``A survey on unknown
  presentation attack detection for fingerprint,'' \emph{arXiv preprint
  arXiv:2005.08337}, 2020.

\bibitem{livdet2009}
G.~M. Marcialis, A.~Lewicke, B.~Tan, P.~Coli, D.~Grimberg \emph{et~al.},
  ``First international fingerprint liveness detection competition - {LivDet}
  2009,'' in \emph{Proc. Intl. Conf. on Image Analysis and Processing
  ({ICIAP})}, 2009, pp. 12--23.

\bibitem{livdet2011}
D.~Yambay, L.~Ghiani, P.~Denti, G.~L. Marcialis, F.~Roli, and S.~Schuckers,
  ``{LivDet} 2011 fingerprint liveness detection competition 2011,'' in
  \emph{Proc. Intl. Conf. on Biometrics (ICB)}.\hskip 1em plus 0.5em minus
  0.4em\relax IEEE, 2012, pp. 208--215.

\bibitem{livdet2015}
V.~Mura, L.~Ghiani, G.~L. Marcialis, F.~Roli, D.~Yambay, and S.~Schuckers,
  ``{LivDet} 2015 fingerprint liveness detection competition 2015,'' in
  \emph{Proc. Intl. Conf. on Biometrics Theory, Applications and Systems
  (BTAS)}.\hskip 1em plus 0.5em minus 0.4em\relax IEEE, 2015, pp. 1--6.

\bibitem{livdet2017}
V.~Mura, G.~Orr{\`u}, R.~Casula, A.~Sibiriu, G.~Loi \emph{et~al.}, ``{LivDet}
  2017 fingerprint liveness detection competition 2017,'' in \emph{Proc. Intl.
  Conf. on Biometrics (ICB)}, 2018.

\bibitem{livdet2019}
G.~Orr{\`u}, R.~Casula, P.~Tuveri, C.~Bazzoni, G.~Dessalvi \emph{et~al.},
  ``{LivDet} in action - fingerprint liveness detection competition 2019,'' in
  \emph{Proc. Intl. Conf. on Biometrics ({ICB})}.\hskip 1em plus 0.5em minus
  0.4em\relax IEEE, 2019, pp. 1--6.

\bibitem{Chugh-FingerprintSpoofBuster-TIFS-2018}
T.~Chugh, K.~Cao, and A.~Jain, ``Fingerprint spoof buster: Use of
  minutiae-centered patches,'' \emph{IEEE Trans. on Information Forensics and
  Security ({TIFS})}, vol.~13, no.~9, pp. 2190--2202, 2018.

\bibitem{livdet2013}
L.~Ghiani, D.~Yambay, V.~Mura, S.~Tocco, G.~L. Marcialis \emph{et~al.},
  ``{LivDet} 2013 fingerprint liveness detection competition 2013,'' in
  \emph{Proc. Intl. Conf. on Biometrics (ICB)}.\hskip 1em plus 0.5em minus
  0.4em\relax IEEE, 2013, pp. 1--6.

\bibitem{DCGANarchitecture}
A.~Radford, L.~Metz, and S.~Chintala, ``Unsupervised representation learning
  with deep convolutional generative adversarial networks,'' \emph{arXiv
  preprint arXiv:1511.06434}, 2015.

\bibitem{Spinoulas-BATLDatasetFingerPAD-Arxiv-2020}
L.~Spinoulas, H.~Mirzaalian, M.~Hussein, and W.~AbdAlmageed, ``Multi-modal
  fingerprint presentation attack detection: Evaluation on a new dataset,''
  \emph{arXiv preprint arXiv:2006.07498}, 2020.

\bibitem{Spinoulas-BATLDesignCaptureDevices-Arxiv-2020}
L.~Spinoulas, M.~Hussein, D.~Geissb{\"u}hler, J.~Mathai, O.~G. Almeida,
  G.~Clivaz, S.~Marcel, and W.~AbdAlmageed, ``Multispectral biometrics system
  framework: Application to presentation attack detection,'' \emph{arXiv
  preprint arXiv:2006.07489}, 2020.

\bibitem{Fritzpatrick-SkinTypes-Dermatology-1988}
T.~B. Fitzpatrick, ``The validity and practicality of sun-reactive skin types
  {I} through {VI},'' \emph{Archives of Dermatology}, vol. 124, no.~6, pp.
  869--871, 1988.

\bibitem{Steiner-facePADswir-ICB-2016}
H.~Steiner, A.~Kolb, and N.~Jung, ``Reliable face anti-spoofing using
  multispectral {SWIR} imaging,'' in \emph{Proc. Intl. Conf. on Biometrics
  (ICB)}, 2016, pp. 1--8.

\bibitem{Senarathna-LSCI-IEEERevBiomedEng-2013}
J.~Senarathna, A.~Rege, N.~Li, and N.~V. Thakor, ``Laser speckle contrast
  imaging: Theory, instrumentation and applications,'' \emph{IEEE Reviews in
  Biomedical Engineering}, vol.~6, pp. 99--110, 2013.

\bibitem{Keilbach-Fingerprint-LSCI-PAD-BIOSIG-2018}
P.~Keilbach, J.~Kolberg, M.~Gomez-Barrero, C.~Busch, and H.~Langweg,
  ``Fingerprint presentation attack detection using laser speckle contrast
  imaging,'' in \emph{Proc. Intl. Conf. of the Biometrics Special Interest
  Group (BIOSIG)}, 2018, pp. 1--6.

\bibitem{Kolberg-LSCIBenchmarkFingerPAD-BIOSIG-2019}
J.~Kolberg, M.~Gomez-Barrero, and C.~Busch, ``Multi-algorithm benchmark for
  fingerprint presentation attack detection with laser speckle contrast
  imaging,'' in \emph{Proc. Intl. Conf. of the Biometrics Special Interest
  Group ({BIOSIG})}, 2019, pp. 1--5.

\bibitem{GomezBarrero-MSSWIRCNN-CRC-2020}
M.~Gomez-Barrero, R.~Tolosana, J.~Kolberg, and C.~Busch, ``Multi-spectral short
  wave infrared sensors and convolutional neural networks for biometric
  presentation attack detection,'' in \emph{AI and Deep Learning in Biometric
  Security: Trends, Potential and Challenges}.\hskip 1em plus 0.5em minus
  0.4em\relax CRC Press, 2020.

\bibitem{Kolberg-LSTM-FingerPAD-IJCB-2020}
J.~Kolberg, A.~C. Vasile, M.~Gomez-Barrero, and C.~Busch, ``Analysing the
  performance of {LSTMs} and {CNNs} on 1310 nm laser data for fingerprint
  presentation attack detection,'' in \emph{Proc. Intl. Joint Conf. on
  Biometrics (IJCB)}, 2020.

\bibitem{lstm}
S.~Hochreiter and J.~Schmidhuber, ``Long short-term memory,'' \emph{Neural
  Computation}, vol.~9, no.~8, pp. 1735--1780, 1997.

\bibitem{lrcn}
J.~Donahue, L.~{Anne Hendricks}, S.~Guadarrama, M.~Rohrbach, S.~Venugopalan
  \emph{et~al.}, ``Long-term recurrent convolutional networks for visual
  recognition and description,'' in \emph{Proc. of the IEEE Conf. on Computer
  Vision and Pattern Recognition (CVPR)}, 2015, pp. 2625--2634.

\bibitem{vgg}
K.~Simonyan and A.~Zisserman, ``Very deep convolutional networks for
  large-scale image recognition,'' \emph{arXiv preprint arXiv:1409.1556}, 2014.

\bibitem{vggface}
O.~M. Parkhi, A.~Vedaldi, and A.~Zisserman, ``Deep face recognition,''
  \emph{British Machine Vision Association}, 2015.

\bibitem{imagenet}
A.~Krizhevsky, I.~Sutskever, and E.~Geoffrey, ``{ImageNet} classification with
  deep convolutional neural networks,'' in \emph{Advances in Neural Information
  Processing Systems 25}.\hskip 1em plus 0.5em minus 0.4em\relax Curran
  Associates, Inc., 2012, pp. 1097--1105.

\bibitem{lfw}
G.~B. Huang, M.~Ramesh, T.~Berg, and E.~Learned-Miller, ``Labeled faces in the
  wild: A database for studying face recognition in unconstrained
  environments,'' University of Massachusetts, Amherst, Tech. Rep. 07-49,
  October 2007.

\bibitem{GomezBarrero-PAD-SWIR-LSCI-ICB-2019}
M.~Gomez-Barrero, J.~Kolberg, and C.~Busch, ``Multi-modal fingerprint
  presentation attack detection: Looking at the surface and the inside,'' in
  \emph{Proc. Intl. Conf. on Biometrics (ICB)}, 2019, pp. 1--8.

\bibitem{He-ResidualDeepLearning-CVPR-2016}
K.~He, X.~Zhang, S.~Ren, and J.~Sun, ``Deep residual learning for image
  recognition,'' in \emph{Proc. of the IEEE Conf. on Computer Vision and
  Pattern Recognition ({CVPR})}, 2016, pp. 770--778.

\bibitem{Szegedy-InceptionV4-AI-2017}
C.~Szegedy, S.~Ioffe, V.~Vanhoucke, and A.~A. Alemi, ``Inception-v4,
  inception-resnet and the impact of residual connections on learning,'' in
  \emph{Proc. of the AAAI Conf. on Artificial Intelligence}, 2017, pp.
  4278--4284.

\bibitem{Tolosana-SWIR-PAD-CNNs-TIFS-2020}
R.~Tolosana, M.~Gomez-Barrero, C.~Busch, and J.~Ortega-Garcia, ``Biometric
  presentation attack detection: Beyond the visible spectrum,'' \emph{IEEE
  Trans. on Information Forensics and Security (TIFS)}, 2020.

\bibitem{mobilenetv2}
M.~Sandler, A.~Howard, M.~Zhu, A.~Zhmoginov, and L.~Chen, ``{MobileNetV2}:
  Inverted residuals and linear bottlenecks,'' in \emph{Proc. of the IEEE Conf.
  on Computer Vision and Pattern Recognition (CVPR)}, 2018, pp. 4510--4520.

\bibitem{Gomez-Barrero-FusionBATL-PAD-UBIO-2018}
M.~Gomez-Barrero, J.~Kolberg, and C.~Busch, ``Towards multi-modal finger
  presentation attack detection,'' in \emph{Proc. Intl. Workshop on Ubiquitous
  implicit BIOmetrics and health signals monitoring for person-centric
  applications (UBIO)}, 2018, pp. 547--552.

\bibitem{Hussein-LSCI-SWIR-CNN-FingerPAD-WIFS-2018}
M.~Hussein, L.~Spinoulas, F.~Xiong, and W.~Abd-Almageed, ``Fingerprint
  presentation attack detection using a novel multi-spectral capture device and
  patch-based convolutional neural networks,'' in \emph{IEEE Workshop on
  Information Forensics and Security (WIFS)}, 2018.

\bibitem{ISO-IEC-30107-3-PAD-metrics-170227}
{ISO/IEC JTC1 SC37 Biometrics}, \emph{{ISO/IEC} 30107-3. Information Technology
  - Biometric presentation attack detection - Part 3: Testing and Reporting},
  2017.

\end{thebibliography}

\end{document}